\documentclass[11pt,letterpaper]{article}

\usepackage{setspace}
\usepackage[letterpaper,margin=1in]{geometry}

\usepackage{graphicx} % Required for inserting images
\usepackage{enumerate}

\usepackage[linesnumbered,ruled,vlined]{algorithm2e}

\usepackage{longtable}
\usepackage{multirow}
\usepackage{amssymb}
\usepackage{float}
\usepackage{upgreek}
\usepackage{appendix}
\usepackage{tabularx}

\usepackage{amsthm}

\usepackage{url}  
\usepackage{hyperref} 

\usepackage[T1]{fontenc}
\usepackage[utf8]{inputenc} 

\newcommand{\Ilt}{\mathcal{I}_{<i}} % = {1,...,i-1}
\newcommand{\Jset}{\mathcal{J}}     % = {1,...,T}

\theoremstyle{remark}

\newtheorem{remark}{Remark}

\usepackage{array, makecell}
\usepackage{longtable}
\usepackage{makecell}

\usepackage{array} % For adjusting row height
\usepackage{amsmath}
\usepackage{graphicx}

\usepackage{subcaption}
\usepackage[margin=1in]{geometry}
\usepackage{enumitem}

\usepackage{tikz}
\usepackage{amsmath}
\usepackage{caption}

\usepackage{geometry}
\usepackage{tikz}

\usepackage{setspace}

\usepackage{tikz}
\usetikzlibrary{shapes.geometric, arrows.meta, positioning, decorations.pathreplacing}

\tikzstyle{process} = [rectangle, minimum width=3.8cm, minimum height=1.1cm, text centered, draw=black, fill=blue!10]
\tikzstyle{decision} = [diamond, minimum width=3.5cm, minimum height=1.2cm, text centered, draw=black, fill=green!10, aspect=2.2]
\tikzstyle{arrow} = [thick,->,>=stealth]
\tikzstyle{startstop} = [process]

\usepackage{amsmath,amssymb}

\usepackage{geometry}
\newtheoremstyle{boldlemma}
  {\topsep}    % Space above
  {\topsep}    % Space below
  {\itshape}   % Body font
  {}           % Indent amount
  {\bfseries}  % Theorem head font (this makes "Lemma" bold)
  {.}          % Punctuation after theorem head
  { }          % Space after theorem head
  {}           % Theorem head spec

\theoremstyle{boldlemma}
\newtheorem{lemma}{Lemma}

\usepackage{booktabs}

\newtheorem{theorem}{Theorem}

\newcommand{\ie}{{\it i.e., }}
\newcommand{\eg}{{\it e.g., }}

\usepackage{natbib}
 \bibpunct[, ]{(}{)}{,}{a}{}{,}%

\usepackage[dvipsnames]{xcolor}

\usepackage{authblk}

\title{Press Start to Charge: Videogaming the Online Centralized Charging Scheduling Problem}

\author[1]{Alireza Ghahtarani}
\author[1]{Martin Cousineau}
\author[2,3]{Amir-massoud Farahmand}
\author[1]{Jorge E. Mendoza}

\affil[1]{Department of Logistics and Operations Management, HEC Montréal}
\affil[2]{Department of Computer Science, University of Toronto}
\affil[3]{Department of Computer Engineering and Software Engineering, Polytechnique Montréal}

\date{}

\begin{document}
%\onehalfspacing
\maketitle
%\usepackage{float}
%\section{Sections}
%Intro: importance of that problem (number of home users, decentralized going to centralized for the grid, studied in the literature in a static way/not dynamic), we solve the problem using gamification, players consist of heuristics and a novel approach (imitation learning)

%Literature review: about the problem, stochastic optimization

%Problem definition (perfect information formulation)

%Game version of the problem

%Agents: heuristics, DAgger, RL

%Computational experiments: instances, results

%Conclusion

%\section{Problem variants}
%\begin{itemize}
%    \item Fixed vs variable power chargers: Can we stack blocks of a same car or not?
%    \item Charging w/ or w/o preemption: Do the blocks need to be contiguous or not?
%    \item Episodic vs continuous learning: Episodic learning is generally easier to train on than continuous learning, since we don't need to set a discounting factor. Also, continuous learning requires a reward at each decision epoch, while episodic learning can only have a reward at the end of the episode. Having only a reward at the end of the episode is somewhat harder, but it can allow us to have a reward that takes into accounts all the variations in power throughout the day. In other words, it can allow us to have an objective function that is more meaningful in practice.
%    \item Allowed to revisit previously scheduled blocks: In this variant, we would have another agent that could rearrange previously scheduled blocks. This would require a way (e.g., colors) to track which blocks go together.
%\end{itemize}

\begin{abstract}
We study the online centralized charging scheduling problem (OCCSP). In this problem, a central authority must decide, in real time, when to charge dynamically arriving electric vehicles (EVs), subject to capacity limits, with the objective of balancing load across a finite planning horizon.
To solve the problem, we first gamify it; that is, we model it as a game where charging blocks are placed within temporal and capacity constraints on a grid. We design heuristic policies, train learning agents with expert demonstrations, and improve them using Dataset Aggregation (DAgger). From a theoretical standpoint, we show that gamification reduces model complexity and yields tighter generalization bounds than vector-based formulations. 
Experiments across multiple EV arrival patterns confirm that gamified learning enhances load balancing. In particular, the image-to-movement model trained with DAgger consistently outperforms heuristic baselines, vector-based approaches, and supervised learning agents, while also demonstrating robustness in sensitivity analyses. These operational gains translate into tangible economic value. In a real-world case study for the Greater Montréal Area (Québec, Canada) using utility cost data, the proposed methods lower system costs by tens of millions of dollars per year over the prevailing practice and show clear potential to delay costly grid upgrades.
\end{abstract}


\section{Introduction}

The global electric vehicle (EV) market is expanding at an extraordinary pace. In 2024, more than 17 million EVs were sold worldwide, over 20\% of new passenger cars, compared to just 2.5\% in 2019 \citep{reuters2025evsales, yale2025evadoption}. Adoption spans nearly all developed regions: Norway leads with over 90\% of new sales, Sweden exceeds 60\%, the Netherlands approaches 30\%, France and the United Kingdom each surpass 25\%, Germany nears 18\%, Singapore has reached 40\%, and China is close to 50\% \citep{iea2025evoutlook}. Canada follows with 12.2\% of new registrations in 2024 \citep{spglobal2024canadaq3ev}.

Widespread EV adoption hinges on major investment in charging infrastructure \citep{nelder2019reducing}, but costs differ sharply between public and home charging. In Canada, expanding public infrastructure is projected to require up to \$47 billion in capital \citep{nrcan2022charginginfrastructure}. By contrast, installing a Level 2 home charger costs \$500-\$7,000 \citep{nicholas2019estimating}; equipping 5 million households by 2040 would total only \$2.5-\$12.5 billion \citep{greenbuilding2025evinstall}, or 6-27\% of the public alternative. Home charging is not only cheaper but also more flexible: whereas public stations demand immediate turnover, residential charging spans the entire period between vehicle arrival and departure, creating opportunities for smart scheduling without burdening users.

While home charging is economically attractive, its uncoordinated usage can jeopardize the stability and efficiency of residential grids. Charging during peak hours can intensify load peaks, causing transformer overloading, voltage fluctuations, overheating, and reduced equipment lifespan, thereby increasing operational costs for utilities \citep{gong2011study, elnozahy2013comprehensive}. Moreover, the cumulative effect of uncoordinated residential charging can strain distribution networks, potentially requiring costly infrastructure upgrades and reducing grid reliability \citep{li2024impact}.

To address these challenges, centralized scheduling has emerged as a promising solution. In this approach, a central authority, such as a utility or aggregator, collects real-time data from connected EVs and coordinates their charging to optimize grid performance \citep{wang2022electric}. By leveraging optimization algorithms, centralized scheduling can flatten demand curves, reduce peak loads, and enhance overall grid stability. Prior studies show that it can significantly reduce simultaneous demand spikes, improve infrastructure efficiency, and provide superior coordination, making it well-suited for large-scale smart charging applications \citep{kang2015centralized, atallah2018optimal, dahiwale2024comprehensive}. Centralized residential control has already been deployed in parts of Europe, \eg through the \citet{dreev2025platform} platform in France and the Intelligent Octopus Go program in the UK \citep{octopus_intelligent_go_2025}, where drivers specify only a departure time and required charge while the provider automatically schedules the charging profile in the background. In Intelligent Octopus Go, all smart-scheduled EV charging is billed at a single low rate, so participants benefit from predictable charging costs without having to worry about intraday price variation.

The online centralized charging scheduling problem (OCCSP) captures the dynamic nature of residential charging. Requests arrive sequentially, each with an energy demand and departure deadline, and must be scheduled in real time under limited capacity, with the objective of balancing load across the planning horizon. \citet{de2017complexity} showed that even simplified variants of the OCCSP are computationally intractable, underscoring the difficulty of designing efficient online algorithms. Unlike offline formulations, where all information is known in advance, the online setting requires adaptive policies that make sequential decisions without perfect knowledge of future arrivals.

Classical optimization approaches struggle with the online nature of the OCCSP. Machine learning (ML) approaches offer an alternative by learning adaptive policies directly from data, but their effectiveness depends heavily on how the problem is represented. Recent work on gamification \citep{kullman2023gamifying} shows that dynamic and stochastic optimization problems can be modeled as video games, with states, actions, and rewards expressed in a visual and interactive form. Gamification is not a learning method in itself but a modeling paradigm that enables both optimization-based and learning-based approaches to be applied more naturally. For EV charging, it provides an intuitive representation of scheduling as a sequential placement game, opening the door to ML agents that can learn effective strategies while still allowing for classical solution techniques.

This paper makes three main contributions. First, we reformulate the OCCSP as a video game and show theoretically that this representation reduces model complexity: convolutional image models scale more favorably than vector-based models as the horizon grows, and predicting a small set of local joystick actions yields tighter generalization guarantees than predicting a high-dimensional scheduling vector. Second, we develop a diverse suite of solution methods: (i) mixed-integer programming benchmarks, including an oracle with perfect information and a rolling-horizon reoptimization policy; (ii) interpretable rule-based heuristics for lightweight real-time scheduling; and (iii) learning-based policies trained from expert demonstrations and improved with Dataset Aggregation (DAgger). Third, we conduct extensive computational experiments across diverse EV arrival patterns and sensitivity settings. Our experiments are grounded in a real-world case from the Greater Montréal Area (GMA) in Québec (Canada), with arrivals and costs calibrated to utility parameters. The results show that the learning-based policies within the gamified framework achieve superior load balancing, with our image-to-movement model trained via DAgger consistently outperforming heuristic, optimization-based, and vector-based approaches. 

Beyond methodological contributions, this study delivers insights directly relevant to the operations management of residential energy systems. By reformulating the OCCSP as a game, we show that centralized schedulers can achieve load balancing by reducing local peak demand and smoothing utilization across the planning horizon, offering a scalable and interpretable mechanism for demand-side management. The gamified representation also reduces computational burden, making real-time coordination feasible under high EV penetration. These findings highlight how gamification can serve as a generalizable framework for online scheduling problems, providing utilities and aggregators with actionable strategies for managing distributed resources under uncertainty; in our GMA case study, translating these operational gains into avoided distribution-capacity costs yields meaningful savings with clear potential to defer grid upgrades.

We organize the remainder of the paper as follows. In Section~\ref{litreview}, we review the relevant literature. In Section~\ref{problemform}, we introduce the formal problem definition of the OCCSP. In Section~\ref{gamification}, we present the different representations of the OCCSP including the gamification. In Section~\ref{valueofgame}, we show the theoretical value of gamification. In Section~\ref{solutionapp}, we describe the solution approaches, including optimization-based benchmarks, rule-based heuristics, and learning-based policies. In Section~\ref{results}, we report the computational results, and in Section~\ref{conclusion}, we offer concluding remarks.

\section{Literature Review}\label{litreview}

The optimization of centralized EV charging has been widely studied under static frameworks that assume complete knowledge of key parameters such as EV arrival and departure times, energy demand, and grid conditions. These deterministic models typically seek to minimize charging costs, flatten peak loads, or improve voltage profiles. For example, \cite{zhou2020scheduling} developed a model that jointly minimizes charging costs and the discomfort experienced by EV owners when charging deviates from their preferred schedules, while \cite{liu2020optimal} addressed infrastructure limitations by optimizing charger utilization under a fixed number of chargers and static grid parameters. Other studies have explored fairness-driven scheduling to enhance grid stability and user satisfaction \citep{aswantara2013centralized} or receding horizon control to reduce transformer overloading \citep{yi2020highly}. Other studies also operate in deterministic settings, but extend toward multi-objective goals. \cite{moeini2014online}, for example, deployed a two-stage framework: first, they minimized EV charging costs, then, in a subsequent phase, addressed grid performance by reducing losses, rescheduling, and incorporating wind generation. \cite{lu2022multi} balanced power losses, voltage profiles, and the state of charge (SoC) of vehicles, while \cite{liu2022electric} tackled both peak-valley load imbalance and user cost under deterministic system conditions. \cite{panda2025multi} used mixed-integer programming to minimize energy losses in active distribution networks. Collectively, these works illustrate the breadth of objectives that can be optimized, though all remain within deterministic formulations.

Other studies address uncertainty directly by modeling key parameters as random variables or as bounded sets. Stochastic programming approaches, such as those of \citet{wang2022electric} and \citet{amin2020review}, treat EV arrival times, energy demand, and electricity prices as random variables. While this provides a more realistic representation, it usually requires extensive scenario generation and the solution of large deterministic equivalents, which limits their use in real-time applications. Robust optimization has also been applied to EV charging. \citet{sun2020robust,kandpal2022robust,sone2024robust,boubaker2025multi} bound uncertainty within predefined sets to safeguard against worst-case outcomes. These methods improve resilience but remain essentially static, since they compute all decisions at once rather than adapting as new information arrives.

To overcome the limits of static and stochastic models, several studies investigate online scheduling for real-time EV charging. Classical policies such as Earliest Deadline First, Least Laxity First, and Optimal Available have been adapted to prioritize charging requests based on deadlines or slack. However, their performance drops under time-varying arrivals \citep{he2012optimal, tang2016online}. Building on competitive analysis, \cite{alinia2018competitive,alinia2020online,alinia2019online} proposed algorithms with provable guarantees under deadline, on-arrival commitment, and global peak constraints. More recent contributions broaden the scope: \cite{sun2020orc} integrated charging with station recommendation in a joint online framework, and \cite{yang2019novel} developed a hierarchical protocol to coordinate large infrastructures. Parallel to this, a structure-driven direction draws insight from offline optimal schedules, especially for convex load objectives, where flattening aggregate demand is known to be optimal \citep{tang2014online, gan2012optimal, tang2016model}. These structural results inspired online strategies that modulate charging to avoid load spikes, even without future information. Together, these works extend classical online scheduling to capture the constraints and scale of real charging networks.

A different line of research explores hybrid approaches that combine elements of centralized and decentralized scheduling. These methods rely on a central coordinator to set system-level objectives, while leaving part of the decision-making to local agents. \cite{nejad2017online} and \cite{bostandoost2023near} designed online algorithms that combine real-time scheduling with dynamic pricing to maximize social welfare under capacity constraints, emphasizing coordination at the station level. \cite{koufakis2019offline} proposed a hybrid mechanism that allows vehicle-to-vehicle energy sharing, extending the range of feasible scheduling options. \cite{wang2024semi} introduced a semi-decentralized framework for parking lots, where local optimization is coupled with global coordination. Most of these studies focus on public charging, where shared capacity and price signals are central. In contrast, research on home charging remains limited; \cite{van2024grid} proposed heuristics for household-level charging under local grid constraints. Taken together, these works underscore the value of coordinated yet distributed control in public settings but leave open the case of centralized home charging at scale, precisely the setting we address.

Finally, recent work applies ML to online EV charging. In the single-vehicle setting, \cite{zhang2020cddpg} casted the problem as a Markov decision process (MDP) and used deep deterministic policy gradient (DDPG) to minimize electricity cost while ensuring the vehicle reaches its required SoC by departure, and \cite{yan2021deep} developed a soft actor-critic framework that enables continuous control and models user anxiety during charging. In station-based contexts, \cite{colak2024deep} employed deep Q-learning (DQN) to allocate fast-charging resources in real time under priority service; \cite{xia2025user} scaled to a network of stations via a two-stage design, load-balancing matching across stations followed by double DQN scheduling within each station, to jointly reduce user costs and balance grid load; and \cite{ding2020optimal} incorporated renewables and vehicle-to-grid capabilities in a multi-spot station, using multi-agent reinforcement learning with DDPG to minimize grid electricity cost while ensuring all EVs meet their SoC targets. Collectively, these studies demonstrate the versatility of ML for handling uncertainty and complex charging environments, but they remain limited to either single-vehicle or public-station settings, leaving large-scale home charging coordination largely unexplored.

While reinforcement learning (RL) is a powerful framework for sequential decision-making, it often requires extensive exploration and large volumes of interaction data, and training can be unstable and sensitive to hyperparameters (see, \eg \citet{henderson2018deep}). When the environment is known or high-quality expert demonstrations are available, RL may also introduce unnecessary complexity. These limitations motivate the use of supervised learning (SL), which leverages expert-labeled data to enable faster and more reliable training. Building on image-centric decision models from the deep RL literature \citep{mnih2015human, tang2020reinforcement, cappart2021combinatorial}, \citet{kullman2023gamifying} recently showed that online decision problems can be framed as interactive games with image-based states and joystick-like actions, allowing learning from visual representations. However, they neither incorporate expert optimization solutions as supervision nor evaluate the benefits of gamification itself. In this work, we pair gamification with expert-driven supervised learning to develop an effective and interpretable approach to centralized EV charging scheduling. Moreover, unlike much of the ML-based charging literature, which focuses on station-level infrastructure or dynamic price signals, we address centralized coordination of home charging across multiple households, where a central scheduler balances aggregate load while meeting individual user needs, a setting of growing importance as EV adoption accelerates and operators seek scalable demand-management mechanisms beyond localized or price-driven strategies.

\section{The OCCSP}\label{problemform}

In centralized EV home charging systems, decisions must be made sequentially as vehicles arrive over time. Future arrivals and energy demands are unknown at decision points, requiring a framework that supports adaptive, real-time decision-making under uncertainty. To model this dynamic setting, we formulate the OCCSP as a finite-horizon MDP. Before introducing the MDP elements, we first define the notion of a \emph{charging block}, which is central to our formulation. 

A \emph{charging block} represents the indivisible task of serving an EV $i$. When vehicle $i$ plugs into its home charger, a new job is created with a specific \emph{length} (the number of charging slots required) and a \emph{power level} (per-slot charging capacity). We will refer to this as the charging block of vehicle $i$ and we hereon assume that all charging blocks have the same power level $u$. Each decision epoch corresponds to the arrival of a new EV and the creation of its associated charging block, which must be scheduled into the system while respecting feasibility constraints.

\subsection{Problem Setting}

We consider a finite planning horizon of $T$ discrete charging time slots. Decisions are made at epochs $i \in \{1,\dots,N\}$, where each epoch corresponds to the arrival of the $i$-th EV and the introduction of its charging block. At that moment, the scheduler must decide how to allocate the block within the available time window. 

The OCCSP is formulated as an MDP defined by the tuple 
$\mathcal{M} = (\mathcal{S}, \mathcal{A}, \mathcal{P}, \mathcal{R})$, where $\mathcal{S}$ is the state space, $\mathcal{A}$ is the action space, $\mathcal{P}$ is the transition function, and $\mathcal{R}$ is the reward function. 

\subsection{State Space}

At epoch $i$, the system is in state $s_i \in \mathcal{S}$. The state is defined as
$s_i = (\mathbf{c}_i, b_i)$,
where $\mathbf{c}_i = \big( c_i(1), c_i(2), \dots, c_i(T) \big) \in \mathbb{N}^T$ represents the current load profile (the number of charging blocks already assigned to each time slot), and $b_i = (ar_i, d_i, l_i)$ encodes the attributes of the charging block for EV $i$. Here, $ar_i, d_i \in \{1,\dots,T\}$ denote the arrival and departure slots of vehicle $i$, respectively, while $l_i$ is the length of the charging block, \ie the number of contiguous slots required:
$l_i = \left\lceil \frac{\mathrm{SoC}_i^d - \mathrm{SoC}_i^{ar}}{u} \right\rceil$,
where $\mathrm{SoC}_i^{ar}$ and $\mathrm{SoC}_i^d$ are the arrival and desired departure states of charge, and $u$ is the energy delivered per slot. 

We assume charging can only start at slot boundaries. When continuous timestamps $(\widetilde{ar}_i, \tilde d_i)$ are observed, we discretize by setting $ar_i := \lceil \widetilde{ar}_i \rceil$ and $d_i := \lfloor \tilde d_i \rfloor$. Thus, each EV’s charging block is defined by the triplet $(ar_i, d_i, l_i)$.

\subsection{Action Space}

Given the current state $s_i$, the scheduler must select an action $a_i \in \mathcal{A}(s_i)$, which represents a feasible charging schedule for the $i$-th EV. An action determines exactly which consecutive time slots within the vehicle’s availability window $[ar_i, d_i]$ will be allocated to satisfy its charging block of length $l_i$. A feasible action must satisfy the following conditions:

\begin{itemize}
    \item In every selected slot, the system’s capacity limit is respected; 
    \item The chosen slots must form a contiguous sequence, since charging cannot be preempted; 
    \item The number of allocated slots must equal the vehicle’s charging requirement $l_i$, and all of them must lie within the interval $[ar_i, d_i]$. 
\end{itemize}

The admissible action set $\mathcal{A}(s_i)$ therefore depends jointly on the current load profile $\mathbf{c}_i$ and the parameters of the charging block $b_i = (ar_i, d_i, l_i)$. Each action fully specifies the charging decision for the current EV and prepares the system to handle the next arrival.

\subsection{State Transitions}

After an action $a_i$ is chosen for the current EV, the system transitions to a new state $s_{i+1} \in \mathcal{S}$ according to the stochastic transition function $\mathcal{P}(s_{i+1} \mid s_i, a_i)$. This transition reflects two key updates. First, the load profile $\mathbf{c}_{i+1}$ is recalculated by increasing the counts of those time slots that have been assigned to the charging block of vehicle $i$. In other words, once the scheduler determines in which slots the EV will charge, the corresponding entries of the load vector are incremented. At this point, the charging request of vehicle $i$ is fully scheduled. The system then proceeds to the next epoch, when a new EV arrives and generates its own charging block $b_{i+1}$, which becomes the next scheduling task. In this way, each transition both incorporates the charging decision for the current vehicle and prepares the system for the uncertainty of future arrivals.

\subsection{Reward Function}

The reward function encourages load balancing by penalizing decisions that exacerbate peak-valley imbalances in the load profile. Define
$\Delta_i = \max_{j \in \{1,\dots,T\}} c_i(j) - \min_{j \in \{1,\dots,T\}} c_i(j)$, as the imbalance before action $a_i$, and 
$\Delta_{i+1} = \max_{j \in \{1,\dots,T\}} c_{i+1}(j) - \min_{j \in \{1,\dots,T\}} c_{i+1}(j)$, as the imbalance afterward. The reward is given by:
\begin{equation}\label{eq:reward_function}
 \mathcal{R}(s_i, a_i) =
\begin{cases}
+1, & \text{if } \Delta_{i+1} < \Delta_i, \\
-1, & \text{if } \Delta_{i+1} > \Delta_i, \\
\phantom{+}0, & \text{otherwise.}
\end{cases}   
\end{equation}
This formulation provides a simple and interpretable learning signal that rewards actions improving load balance and penalizes those that worsen it.

\subsection{Objective}

The objective in this MDP framework is to identify a policy $\pi: \mathcal{S} \to \mathcal{A}$ that maximizes the expected cumulative reward over the planning horizon:
$\max_{\pi} \; \mathbb{E}\left[ \sum_{i=1}^{N} \mathcal{R}(s_i, \pi(s_i)) \right]$.
This enables the design of intelligent, adaptive policies that can make effective charging decisions in real time while accounting for uncertainty in EV arrivals and the dynamic evolution of the system.

\section{Representation of the Problem}\label{gamification}

Having formulated the OCCSP as an MDP in Section~\ref{problemform}, the next step is to determine how this abstract formulation should be represented in practice. The effectiveness of any solution approach, whether based on machine learning or optimization, depends critically on how the state, action, and outcome spaces are encoded. A suitable representation must preserve the essential constraints of the problem while also enabling efficient computation and learning. Different representations emphasize different aspects of the OCCSP, such as spatial visualization, engineered features, or direct scheduling decisions, and therefore lead to distinct trade-offs in complexity, interpretability, and performance. In the following subsections, we present several alternative representations, each designed to bridge the gap between the theoretical MDP and practical solution methods.

\subsection{Vector-based Representation of the OCCSP}\label{vectorrep}

In the vector-based representation, the problem is encoded as a fixed-length vector. This vector may include any set of features extracted from the state of the problem. One natural representation is a vector of $L+T$, which has two components: the first part is a binary vector of length \(L\), where \(L\) denotes the maximum possible length of the charging block. For an incoming EV with length of the charging block \(l\), the first \(l\) entries are set to 1 and the remaining \(L-l\) entries are set to 0, thereby encoding the length of the charging block. The second part is an integer vector of length \(T\), where \(T\) represents the number of discrete time slots, and each entry records the number of EVs already scheduled in the corresponding slot. Together, these two components yield a structured input of dimension \(L+T\) that captures both the requirements of the incoming EV and the current load profile of the system.

The output in this representation is also vector-based. Specifically, it is a binary vector of size \( T \), where a single entry equal to 1 indicates the chosen starting time for the EV’s charging, and all other entries are set to 0. This output directly specifies the scheduling decision by identifying the first slot allocated to the block.

\subsection{Videogaming the OCCSP}\label{gamerep}

Another way to represent the OCCSP is through gamification, in which the scheduling task is reformulated as an interactive game environment. This approach preserves the core structural constraints of the original optimization model but recasts the decision-making process into a real-time, spatially visualized format. The resulting game environment represents the system state as an image and defines a small set of discrete actions, enabling the efficient use of ML techniques. Importantly, the objective of the game remains aligned with that of the original model: achieving load balancing across discrete time slots by strategically allocating EV charging requests within a finite capacity horizon.

\begin{figure}[H]
    \centering
    \includegraphics[width=\textwidth]{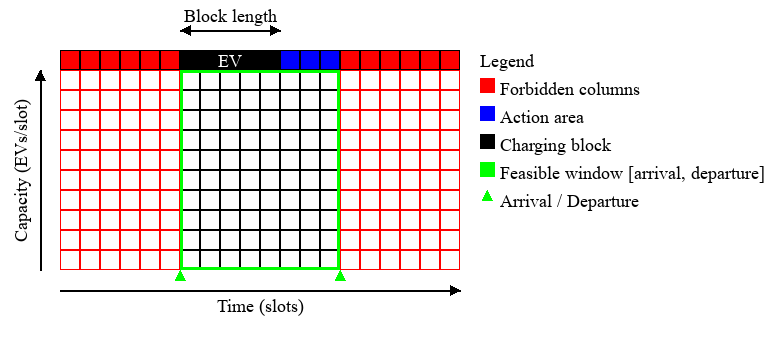}
    \caption{Visualization of the EV scheduling game environment}
    \label{fig:ev_game_board}
\end{figure}

The OCCSP game (Figure \ref{fig:ev_game_board}) is represented as a two-dimensional grid, where the horizontal dimension enumerates discrete time slots and the vertical dimension enumerates capacity units, each cell thus representing a potential allocation of one charging unit at one time step. Players interact with the game through an action area located at the top of the grid, where each block symbolizes the charging demand of an individual EV. The width of a block represents the vehicle’s required charging duration. For each EV, only a subset of columns, those lying between its arrival and departure times, constitutes its feasible placement region on the grid; all other columns are marked as forbidden, enforcing temporal feasibility. Within this region, players can move blocks horizontally and drop them onto the grid. Each block descends as a whole until any part of it contacts the bottom of the grid or the upper boundary of another block, forming a Tetris-like stacking pattern that displays the evolution of system utilization over time. The game’s scoring mechanism, defined by Equation \ref{eq:reward_function}, assigns rewards that encourage actions minimizing the peak-to-valley load difference.

To reflect realistic decision-making constraints, the game enforces a limit on the number of actions a player can take while maneuvering a block. Specifically, if a player exceeds a predefined number of actions, equal to the width of the game environment, which represents the theoretical maximum required to explore all feasible positions, the block is automatically dropped in its current location. Since blocks appear based on their arrival time and earlier time slots are not allowed for scheduling, moving forward allows the player to find the appropriate start time for scheduling. In practice, the width of the game environment provides enough actions to determine the optimal scheduling start time without the risk of permanent loops.

\subsection{Combinations of Inputs and Outputs}\label{sec:combinations}

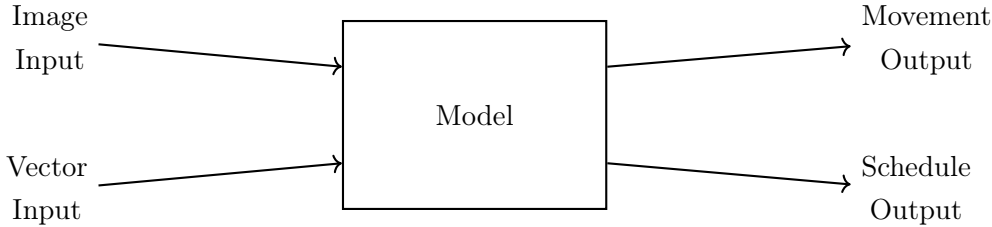
\begin{figure}[H]
    \centering
    \begin{tikzpicture}[thick, node distance=2cm, every node/.style={align=center}]
        % Input nodes
        \node (img) [left=3cm] at (0,1) {Image\\Input};
        \node (vec) [left=3cm] at (0,-1) {Vector\\Input};

        % Model node
        \node (model) [draw, rectangle, minimum width=3.5cm, minimum height=2.5cm] at (2,0) {Model};

        % Output nodes
        \node (act) [right=3cm] at (4,1) {Movement\\Output};
        \node (vecout) [right=3cm] at (4,-1) {Schedule\\Output};

        % Arrows
        \draw[->] (img) -- (model.160);
        \draw[->] (vec) -- (model.200);
        \draw[->] (model.20) -- (act);
        \draw[->] (model.-20) -- (vecout);
    \end{tikzpicture}
    \caption{Simplified illustration of model variants based on input and output types}
    \label{fig:model_variants}
\end{figure}

The OCCSP can be formulated in several ways depending on how the system state and the scheduling decision are represented. In this work, we focus on four representative input-output modalities that combine two types of state representations, vector-based and image-based, with two types of decision outputs, scheduling and movement actions. The vector representation (Section~\ref{vectorrep}) encodes the system as a $L+T$ vector, while the image representation (Section~\ref{gamerep}) expresses the same information as a grid depicting available capacity. A scheduling output is defined as a binary vector of length $T$, where a single entry equal to 1 indicates the chosen starting time for an EV’s charging session. In contrast, a movement output corresponds to one of the discrete actions {\texttt{left}, \texttt{right}, \texttt{down}}, which determine how a charging block is positioned on the grid. Combining these two dimensions yields four variants, Vector-to-Schedule (V2S), Vector-to-Movement (V2M), Image-to-Schedule (I2S), and Image-to-Movement (I2M), illustrated in Figure~\ref{fig:model_variants}. Although other feature augmentations are possible (\eg including arrival/departure windows or EV attributes), all formulations share the same underlying temporal structure and thus satisfy the theoretical properties established in the next sections.

These four combinations, illustrated in Figure~\ref{fig:model_variants}, define the design space for representing the OCCSP within learning frameworks. Each highlights different structural aspects of the problem and leads to distinct trade-offs in complexity, interpretability, and computational efficiency.

\section{The Theoretical Value of Gamification}\label{valueofgame}

Having introduced the four possible input, output combinations (see Figure~\ref{fig:model_variants}), I2M, I2S, V2M, and V2S, we now turn to their theoretical comparison. The goal is to assess the value of visual inputs, movement-based outputs, and the sequential dynamics that arise through gamification. Among these variants, the I2M formulation represents the game-based version and serves as a natural baseline for evaluating whether the structural simplicity of vector-based models can achieve performance comparable to the flexibility and realism of an interactive environment.

To theoretically compare the four models, we conduct a formal complexity analyses grounded in statistical learning theory across both input and output modalities. For input modalities, image-based versus vector-based, the architectural choice differs: we assume convolutional neural networks (CNNs) for images and fully connected networks (MLPs) for vectors. This difference necessitates an asymptotic comparison based on parameter growth and generalization bounds (see Lemma~\ref{lem:ev-param-gap} and Theorem~\ref{theory1}). In contrast, the comparison of output types, 3-movement prediction versus $T$-dimensional vector prediction, can be carried out within the same model class, allowing a non-asymptotic analysis of sample complexity under equal network capacity (see Theorem~\ref{thm:MvT-complete}). The Vapnik-Chervonenkis (VC) dimension, defined as the largest number of points a hypothesis class can label in all possible ways, serves as our measure of model complexity. We apply it in Lemma~\ref{lem:ev-param-gap} to compare CNN- and MLP-based architectures, with the resulting differences in VC dimension leading to distinct generalization bounds in Theorem~\ref{theory1}. Proofs of Lemma~\ref{lem:ev-param-gap} and Theorems~\ref{theory1} and \ref{thm:MvT-complete} are provided in Appendix~\ref{sec:appendixA}.

\begin{lemma}[CNN VC term vs.\ MLP parameter term]\label{lem:ev-param-gap}
Let us consider a CNN that uses global pooling with $C$ convolution layers whose architectural hyperparameters (number of filters, kernel sizes, strides, paddings, and pooling windows) are independent of $T$. Let $\mathcal H_I$ denote the set of all policies that can be implemented by such CNNs with image input and rectified linear units (ReLUs) as the activation functions (\ie the CNN hypothesis class), and let $\Phi_I$ denote an upper bound on the VC dimension $\mathrm{VCdim}(\mathcal H_I)$. Similarly, let $\mathcal H_V$ denote the set of all policies that can be implemented by an MLP (\ie the MLP hypothesis class), with vector input, and let $W_V$ be the MLP parameter count for a vector input of length $d=L+T$ and hidden width $\eta$ (with $L, \eta, |\mathcal A|$ fixed w.r.t.\ $T$). Then, as $T\to\infty$,
\[
\Phi_I \;=\; \Theta\!\big(\log(C\,\lambda)\big)
\qquad\text{while}\qquad
\Phi_V \;=\; W_V\log W_V \;=\; \Theta\!\big(T\log T\big).
\]
In particular, if the image resolution $\lambda$ (the width of the input image in pixels, \ie the number of pixel columns representing the $T$ time slots) scales linearly with $T$ (\ie $\lambda\asymp T$), then $\Phi_I=\Theta(\log T)$ and hence $\Phi_I \ll \Phi_V$.

\end{lemma}

Based on the Lemma \ref{lem:ev-param-gap}, the MLP’s size scales \emph{linearly} with the time‐grid resolution $T$, due to the $(L+T)\,m$ term. The CNN’s size grows only logarithmically in $T$ (as in Lemma~\ref{lem:ev-param-gap}) and relies only on the fixed filter count $F$ and spatial size $k$. As the discretization is refined ($T$ increases), the CNN remains compact whereas the MLP grows without bound, yielding a vanishing parameter‐ratio.

\begin{theorem}[Comparison of Image vs. Vector Input Estimation Error Bounds]\label{theory1}
Using the VC-dimension growth rates established in Lemma~\ref{lem:ev-param-gap}, the upper bound on the estimation error for image-input CNNs grows as a function of $T$ strictly slower than that for vector-input MLPs.
\end{theorem}

Theorem~\ref{theory1} implies that the estimation error of a hypothesis can be bounded by its empirical (training) error plus a complexity term that depends on the VC dimension of the model class. For vector-input MLPs, this complexity term grows like \(\sqrt{\frac{T \log T}{n}}\), whereas for image-input CNNs grows only like \(\sqrt{\frac{\log T}{n}}\). Therefore, for the same sample size and confidence level, the image-based model has a smaller complexity contribution given condition of Theorem \ref{theory1}, and asymptotically, in $T$, the image-input formulation is preferable to the vector-input one because its term increases only with \(\log T\), not with \(T \log T\).

We now focus on the output representations, movement-based versus schedule-vector, illustrated in Figure~\ref{fig:model_variants}. First, using Theorem~\ref{thm:MvT-complete}, we establish the conditions under which the movement-based output yields a tighter generalization guarantee than the schedule-vector output, as captured by the Natarajan bound \citep{natarajan1989learning}. Then, based on Remark~\ref{thm:3v96-revised}, we demonstrate that, in our proposed game setting, the movement output provides a tighter Natarajan bound than the schedule-vector representation.

\begin{theorem}[Comparison of \(M\)-class vs.\ \(S\)-class generalization terms]
\label{thm:MvT-complete}
Let \(n\) be the number of i.i.d.\ training examples, \(\delta\in(0,1)\) with \(B=\ln(\frac{1}{\delta})\), and write \(X=\ln n+\ln H\).  Suppose the single‐shot “vector” model has \(S\) output classes and Natarajan dimension \(d_S\), while the sequential “action” model makes up to \(H\) decisions each among \(M\) classes and has Natarajan dimension \(d_M\).  Denote by
\[
G_1=\frac{d_S(\ln n+\ln S)+B}{n}
\quad\text{and}\quad
G_2=\frac{d_M(X+\ln M)+B}{n\,H}
\]

Consider the complexity terms appearing in the generalization bounds for the vector and movement models:

\[
  \text{If }
  \frac{d_M}{d_S}
  \;<\;
  \frac{H}{1 + \dfrac{\ln M}{\ln n + \ln S}},
  \quad\text{then}\quad
  G_2 < G_1,
\]

\noindent \ie under this condition the \(M\)-class action model carries a strictly smaller generalization bound than the \(S\)-class vector model.
\end{theorem}

By Theorem~\ref{thm:MvT-complete}, we can \emph{have} our network so that the ratio of Natarajan dimensions
\(\tfrac{d_M}{d_S}\) meets the required bound
\[
  \frac{d_M}{d_S}
  \;<\;
  \frac{H}{1 + \frac{\ln M}{\ln n + \ln S}},
\]
and thus guarantees \(G_2<G_1\), \ie the action‐by‐action model generalizes strictly better than the vector model.

%\paragraph{Non-i.i.d.\ note (effective sample sizes).}
If the data are not i.i.d., replace \(n\) by \(n_{\mathrm{eff}}\in(0,n]\) (effective number of independent episodes) and \(H\) by \(H_{\mathrm{eff}}\in[1,H]\) (effective number of independent decisions per episode), and set \(X_{\mathrm{eff}}=\ln n_{\mathrm{eff}}+\ln H_{\mathrm{eff}}\).
The complexity terms become:
\[
G_1=\frac{d_S(\ln n_{\mathrm{eff}}+\ln S)+B}{n_{\mathrm{eff}}},\qquad
G_2=\frac{d_M\big(X_{\mathrm{eff}}+\ln M\big)+B}{n_{\mathrm{eff}}\,H_{\mathrm{eff}}}.
\]
Consequently, the comparison condition specializes to
\[
  \text{If }
  \frac{d_M}{d_S}
  \;<\;
  \frac{H_{\mathrm{eff}}}{1 + \dfrac{\ln M}{\ln n_{\mathrm{eff}} + \ln S}},
  \quad\text{then}\quad
  G_2 < G_1.
\]

\begin{remark}[Special case for our EV charging game]\label{thm:3v96-revised}
Let \(n = 200\) be the number of demonstration blocks, \(S = 96\) the number of 15-minute slots over a 24-hour horizon, \(M = 3\) the three movement actions (left, right, down), and \(H\) for the average number of actions per block.
  Then
\[
  \ln n + \ln S = \ln(200) + \ln(96)\approx 9.86,
  \quad
  \ln 3 \approx 1.10,
\]
so the threshold in Theorem~\ref{thm:MvT-complete} becomes
\[
  \frac{H}{1 + \tfrac{\ln3}{\ln200 + \ln96}}
  = \frac{H}{1 + \tfrac{1.10}{9.86}}
  \approx \frac{H}{1.11}
  \approx 0.90\,H.
\]
Hence the condition specializes to
\[
  \frac{d_3}{d_{96}} < 0.90\,H.
\]

Worth noting, this comparison assumes that the two models share the same architecture and input representation, differing only in their output layers. In particular, we may think of them as either CNNs or MLPs with identical inputs but distinct output heads. Hence both models share the same “backbone” of parameters \(W_{\rm base}\) and last‐hidden‐layer width \(D\), differing only in the final linear head:
\[
  W_{3} = W_{\rm base} + 3D + 3,
  \quad
  W_{96} = W_{\rm base} + 96D + 96.
\]
Using the MLP nearly‐tight bound \(d_i = O(W_i\ln W_i)\) of Lemma \ref{lem:ev-param-gap}, we estimate:
\[
  \frac{d_3}{d_{96}} \approx \frac{W_{3}\ln W_{3}}{W_{96}\ln W_{96}}.
\]
Since \(W_{3}<W_{96}\), this ratio is strictly less than 1, and as long as
\[
  1 < 0.90\,H
  \quad\Longleftrightarrow\quad
  H > \frac{1}{0.90}\approx1.11,
\]
the inequality \(\tfrac{d_3}{d_{96}}<0.90\,H\) holds automatically.  In practice, with even a modest average \(H\ge2\) moves per block, the 3‐class movement model \emph{provably} enjoys a tighter generalization bound than the 96‐class vector model.

In the CNN setting, both the movement and vector policies share the same image encoding and convolutional backbone; they differ only in the final linear head. Let $\mathcal H_{I,3}$ and $\mathcal H_{I,96}$ denote the corresponding CNN hypothesis classes with 3 movement actions and 96 schedule classes, and let $d_3$ and $d_{96}$ be their Natarajan dimensions as in Remark~\ref{thm:3v96-revised}. After global pooling, the backbone maps each image to a feature vector in $\mathbb{R}^D$, and a linear head with $K$ outputs has $K(D+1)$ parameters. With a fixed backbone (fixed $D$), the complexity contributed by the head therefore scales roughly linearly in $K$, yielding the crude approximation $\frac{d_3}{d_{96}} \;\approx\; \frac{3}{96} \;\approx\; 0.03$, which again indicates that the 3-action movement CNN is substantially less complex than the 96-class vector CNN.

\end{remark}

\begin{remark}[Max-dependence, non-i.i.d]\label{rem:worst-neff}
In the extremal dependence case, take \(n_{\mathrm{eff}}=1\) (maximal cross-episode dependence) and \(H_{\mathrm{eff}}=1\) (maximal within-episode dependence). With \(S=96\) and \(M=3\), the dependent-data threshold becomes
\[
\frac{H_{\mathrm{eff}}\big(\ln n_{\mathrm{eff}}+\ln 96\big)}{\ln n_{\mathrm{eff}}+\ln H_{\mathrm{eff}}+\ln 3}
\;=\;
\frac{\ln 96}{\ln 3}
\;\approx\;4.15.
\]
If both models share the same backbone so that \(d_3<d_{96}\) (e.g., \(d_i\approx W_i\ln W_i\) with \(W_3<W_{96}\)), then
\[
\frac{d_3}{d_{96}} \;<\; 1 \;<\; 4.15,
\]
and thus the comparison condition holds even under this maximal-dependence scenario.
\end{remark}

Combining the parameter-count analysis of Lemma~\ref{lem:ev-param-gap} with the generalization-bound separation in Theorem~\ref{theory1} reveals that any image-based model (I2M or I2S) enjoys a strictly tighter sample-complexity bound than its vector-based counterpart (V2M or V2S) as the number of time slots increases. Furthermore, Theorem~\ref{thm:MvT-complete} establishes, in general form, the condition under which predicting movement actions leads to a smaller Natarajan-dimension complexity term than predicting a full vector. Remark~\ref{thm:3v96-revised} further specializes this result to the proposed game, demonstrating that using three movement actions consistently yields a tighter bound than vector prediction. Together, these results establish that the I2M design, combining an image input with a 3‐way movement output, achieves the most favorable theoretical guarantees among the four input-output combinations in Figure \ref{fig:model_variants}.

\section{Solution Approaches}\label{solutionapp}

In this section, we propose a set of solution approaches to address the OCCSP. These approaches fall into three main categories: methods based on a static formulation of the problem, heuristic strategies that offer fast and interpretable rules for real-time decision-making, and SL-based techniques that learn adaptive policies within the gamified environment to handle uncertainty and sequential arrivals.

\subsection{Solution Approaches Based on the Static Problem}

Two solution methods are grounded in the static formulation of the OCCSP: one assumes full knowledge of all EV arrivals and energy requirements in advance, while the other applies the static model iteratively to each arriving EV. The static setting considers the same notation and parameters introduced in Section~\ref{problemform} for the MDP, with \( \mathcal{I} = \{1, \dots, N\} \) denoting the set of EVs and \( \mathcal{T} = \{1, \dots, T\} \) the set of discrete time slots. For each EV \( i \in \mathcal{I} \), the arrival and departure are given by discrete slot indices \( ar_i, d_i \in \mathcal{T} \), and the charging requirement \( l_i \) is computed as in Section~\ref{problemform}. When continuous timestamps are provided, we use the same rounding-to-slots convention as in Section~\ref{problemform}.
The binary decision variable \( x_{ij} \) indicates whether EV \( i \) is charging in time slot \( j \), and \( z_{ij} \) indicates whether EV \( i \) starts charging in time slot \( j \). The aggregate load in time slot \( j \) is \( p_j = \sum_{i \in \mathcal{I}} x_{ij} \), with \( p^{\max} = \max_{j \in \mathcal{T}} p_j \) and \( p^{\min} = \min_{j \in \mathcal{T}} p_j \). The static formulation assigns each EV a contiguous sequence of \( l_i \) charging time slots within \([a_i, d_i]\), subject to capacity constraints \( p_j \le p^{\mathrm{cap}} \) and load-balancing objectives that minimize the imbalance \( p^{\max} - p^{\min} \). The complete mathematical formulation is given in Model~\eqref{model_1}.

\begin{subequations}\label{model_1}
\begin{alignat}{3}
& F(\mathrm{x},\mathrm{z}) =\min {p^{max}-p^{min}},\label{model_1_const_1} \\
&\text{s.t.}\;\;\; p^{max}\geq p_j \,& \forall j \in\mathcal{T}\label{model_1_const_2}\\
& p^{min}\leq p_j \,& \forall j \in\mathcal{T}\label{model_1_const_3}\\
& p_{j}=\sum_{i=1}^{N}{x_{ij}},&\forall j \in \mathcal{J}\label{model_1_const_4}\\
& p_{j}\leq p^{cap},&\forall j \in \mathcal{J}\label{model_1_const_5}\\
& x_{ij'}=0,&\forall i \in \mathcal{I}, j'\in \mathcal{J}| j'< ar_i \label{model_1_const_6}\\
& x_{ij'}=0,&\forall i \in \mathcal{I}, j'\in \mathcal{J}| j'> d_i \label{model_1_const_7}\\
& \sum_{j=ar_i}^{d_i-l_i}{z_{ij}}=1,&\forall i \in \mathcal{I} \label{model_1_const_8}\\
& z_{ij'}=0,&\forall i \in \mathcal{I}, \forall j'\in \mathcal{J}|j'< ar_i\label{model_1_const_9}\\
& z_{ij'}=0,&\forall i \in \mathcal{I}, \forall j'\in \mathcal{J}|j'> d_i\label{model_1_const_10}\\
& (1-z_{ij})j \geq \sum_{j'\in \mathcal{J}|j'<j}{x_{ij'}},&\forall i \in \mathcal{I}, j \in \mathcal{J} \label{model_1_const_11}\\
& z_{ij} \geq x_{ij}-x_{i,j-1},&\forall i \in \mathcal{I},j\in \mathcal{J} \setminus \{1\} \label{model_1_const_12}\\
& z_{i1}\geq x_{i1}, \label{model_1_const_13}\\
& \sum_{j}x_{ij}=l_i, &\forall i \in \mathcal{I}, \label{model_1_const_14}\\
& x_{ij} \in \{0,1\}, &\forall i \in \mathcal{I}, j \in \mathcal{J}, \label{model_1_const_15}\\
& z_{ij} \in \{0,1\}, &\forall i \in \mathcal{I}, j \in \mathcal{J}. \label{model_1_const_16}
\end{alignat}
\end{subequations}

The objective function ~\eqref{model_1_const_1} minimizes the difference between the maximum and minimum number of EVs scheduled for charging across all time slots. 
Constraints~\eqref{model_1_const_2} and~\eqref{model_1_const_3} define the maximum and minimum number of EVs charged in any time slot, while constraints~\eqref{model_1_const_4} computes the actual load in each time slot as the sum of individual EV charging decisions. Constraints~\eqref{model_1_const_5} enforces the capacity limit \( p^{cap} \) for each time time slot, ensuring that the number of EVs charged does not exceed the system’s physical constraints.
Temporal feasibility is enforced through constraints~\eqref{model_1_const_6} and~\eqref{model_1_const_7}, which prevent EVs from being charged outside their arrival and departure windows. Constraint~\eqref{model_1_const_8} ensures that each EV has a single starting time slot for its charging session, while constraints~\eqref{model_1_const_9} and~\eqref{model_1_const_10} ensure that this starting time slot falls within the allowed window. Constraints~\eqref{model_1_const_11}--\eqref{model_1_const_13} enforce that charging only occurs after the start time slot, preserving contiguity of the charging sequence. Constraints~\eqref{model_1_const_14} guarantees that each EV receives exactly \( l_i \) units of charging time, thereby fully satisfying its energy need within the allowed interval. Finally, constraints \eqref{model_1_const_15} and \eqref{model_1_const_16} enforce the binary constraints.

\subsubsection{Oracle Solution Approach}

We begin with the oracle solution, which assumes complete foresight of all future EV arrivals, charging requirements, and availability windows. In this fully offline setting, the entire sequence of charging requests is known in advance, allowing the scheduler to solve the global optimization problem \eqref{model_1} over the full planning horizon and compute the assignment that minimizes load imbalance while satisfying all constraints.
Although this perfect-information policy is not implementable in practice, it serves as an ideal benchmark. The oracle relaxes the information constraints of the online problem. As shown by \cite{brown2014information}, such information relaxations can only (weakly) improve the objective. Consequently, the oracle’s value provides a fundamental bound on the performance of any nonanticipative (online) strategy, a lower bound for cost-minimization problems and an upper bound for reward-maximization ones. Thus, even though it cannot be executed in real time, the oracle solution offers a gold-standard reference against which practical online policies can be evaluated.

\subsubsection{Re-optimization Solution Approach}\label{reoprtmodel}

Re-optimization is performed by solving Model~\eqref{model_1} repeatedly, once per arrival. At each arrival, all previously made assignments are fixed and only the variables of the newcomer are optimized; EVs that have not yet arrived are excluded from the epoch’s problem. Let \(\bar x_{\iota j},\bar z_{\iota j}\in\{0,1\} \quad \forall\, \iota\in\Ilt,\ \forall\, j\in\Jset\) denote the incumbent decisions from earlier arrivals. The per-arrival model is exactly Model \eqref{model_1} augmented with the fixing equalities:

\begin{equation}\label{eq:fixed-past}
x_{\iota j}=\bar{x}_{\iota j},\ \ z_{\iota j}=\bar{z}_{\iota j}
\quad \forall\, \iota\in\Ilt,\ \forall\, j\in\Jset
\end{equation}

With these equalities in place, committed load naturally enters \(p_j\), \(p^{max}\), \(p^{min}\), and the capacity constraints, and only the arriving EV is decided; no rescheduling is allowed.

\subsection{Heuristic Approaches for Online EV Scheduling}

Heuristic methods offer lightweight, interpretable strategies for making real-time decisions in the OCCSP. By relying on local or myopic information, these methods provide feasible solutions without requiring future knowledge. Although heuristics do not guarantee optimality, they serve as valuable baselines and practical tools, especially in constrained operational environments. In this subsection, we describe three heuristic policies designed to balance load and maintain feasibility.

\subsubsection{Heuristic 1: Row-filling}

The row-filling heuristic mimics a Tetris-like process in which charging blocks are sequentially placed into a load matrix that represents time and capacity. 

At each arrival, the incoming EV's charging block of length~$l_i$ must be inserted within its feasible window~$[ar_i, d_i]$ without exceeding the per-time slot capacity~$p_{\text{cap}}$. The heuristic scans the load matrix from top to bottom (rows) and left to right (columns), identifying the earliest feasible start position that maintains contiguity and respects the capacity constraint. Once the block is placed, the corresponding cells of the matrix are filled, updating the aggregate load before the next arrival.

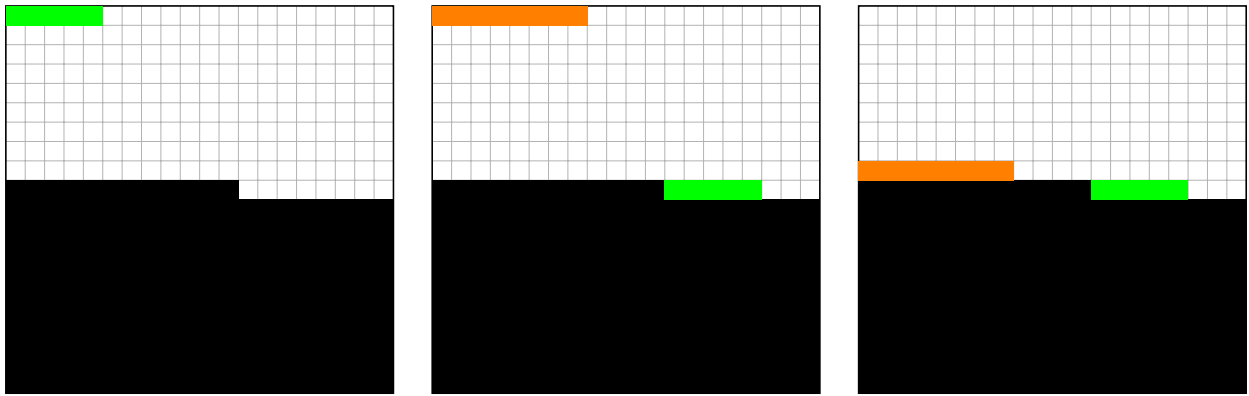
\begin{figure}[H]
\centering
\resizebox{\textwidth}{!}{%
\begin{tikzpicture}[scale=0.35]

  %======================
  % PANEL 1
  %======================
  \begin{scope}[shift={(0,0)}]
    \draw[gray!60] (0,0) grid (20,20);
    \draw[thick] (0,0) rectangle (20,20);

    % Half fill: bottom 10 rows (y=0..9), all columns (x=0..19)
    \foreach \x in {0,...,19}{
      \foreach \y in {0,...,9}{
        \fill (\x,\y) rectangle ++(1,1);
      }
    }

    \foreach \x in {0,...,11}{
      \fill (\x,10) rectangle ++(1,1);
     }
     %\end{tikzpicture}

    \foreach \x in {0,...,4}{
      \fill[green] (\x,19) rectangle ++(1,1);
     }

    % --- EXTRA CELLS (Panel 1): add more black squares here if needed ---
    % \fill (2,12) rectangle ++(1,1);
    % \fill (7,15) rectangle ++(1,1);
  \end{scope}

  %======================
  % PANEL 2 (shifted right by 22 units → 2-unit gap)
  %======================
  \begin{scope}[shift={(22,0)}]
    \draw[gray!60] (0,0) grid (20,20);
    \draw[thick] (0,0) rectangle (20,20);

    \foreach \x in {0,...,19}{
      \foreach \y in {0,...,9}{
        \fill (\x,\y) rectangle ++(1,1);
      }
    }

    \foreach \x in {0,...,11}{
      \fill (\x,10) rectangle ++(1,1);
     }

    \foreach \x in {12,...,16}{
      \fill[green] (\x,10) rectangle ++(1,1);
     }

    \foreach \x in {0,...,7}{
      \fill[orange] (\x,19) rectangle ++(1,1);
     }    

    % --- EXTRA CELLS (Panel 2) ---
    % \fill (3,12) rectangle ++(1,1);
    % \fill (10,13) rectangle ++(1,1);
  \end{scope}

  %======================
  % PANEL 3 (shifted right again by 22 units)
  %======================
  \begin{scope}[shift={(44,0)}]
    \draw[gray!60] (0,0) grid (20,20);
    \draw[thick] (0,0) rectangle (20,20);

    \foreach \x in {0,...,19}{
      \foreach \y in {0,...,9}{
        \fill (\x,\y) rectangle ++(1,1);
      }
    }

    \foreach \x in {0,...,11}{
      \fill (\x,10) rectangle ++(1,1);
     }

    \foreach \x in {12,...,16}{
      \fill[green] (\x,10) rectangle ++(1,1);
     }

    \foreach \x in {0,...,7}{
      \fill[orange] (\x,11) rectangle ++(1,1);
     }

    % --- EXTRA CELLS (Panel 3) ---
    % \fill (5,12) rectangle ++(1,1);
    % \fill (15,14) rectangle ++(1,1);
  \end{scope}

\end{tikzpicture}%
}
%\caption{Row-Filling heuristic.}
\caption{Row-filling heuristic}\label{fig:row-filling-heuristic}
\end{figure}

This ``row-first'' filling strategy has a natural operational interpretation: it schedules new arrivals in the shallowest available regions of the load profile, thereby minimizing vertical stacking and reducing the peak load. The algorithm runs in~$\mathcal{O}(T)$ time per EV, making it suitable for real-time centralized coordination. This heuristic is illustrated in Figure \ref{fig:row-filling-heuristic}.

\subsubsection{Heuristics 2 \& 3: X-threshold}

Row-filling schedules every EV as early as possible, but it does so without considering how congested later time slots may become. As a result, it can push demand into later time slots that end up heavily loaded, even when the current time slot still has room to absorb part of that load. The X-threshold heuristic introduces a simple control at arrival times: it limits how many EVs are assigned immediately upon arrival, encouraging the use of nearby slack while preventing unnecessary peaks at the current time slot.

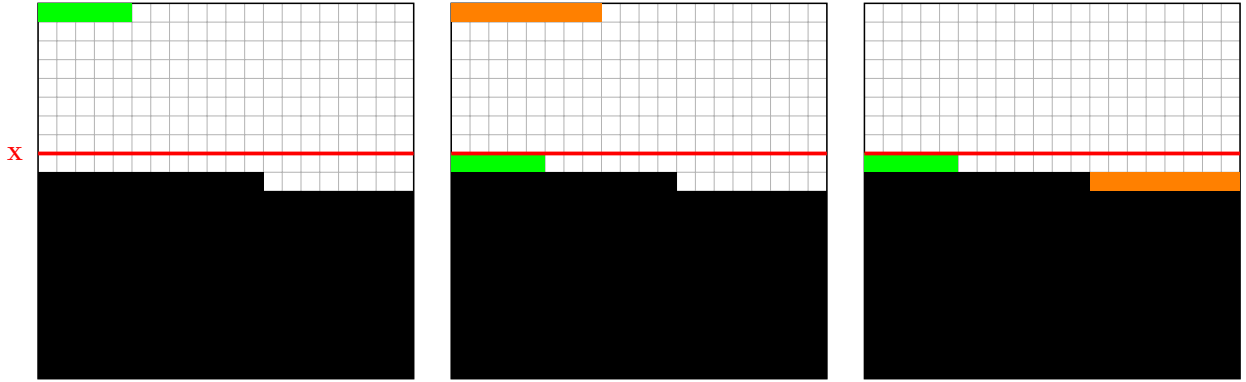
\begin{figure}[H]
\centering
\resizebox{\textwidth}{!}{%
\begin{tikzpicture}[scale=0.35]

  %======================
  % PANEL 1
  %======================
  \begin{scope}[shift={(0,0)}]
    \draw[gray!60] (0,0) grid (20,20);
    \draw[thick] (0,0) rectangle (20,20);

    % Half fill: bottom 10 rows (y=0..9), all columns (x=0..19)
    \foreach \x in {0,...,19}{
      \foreach \y in {0,...,9}{
        \fill (\x,\y) rectangle ++(1,1);
      }
    }

    \foreach \x in {0,...,11}{
      \fill (\x,10) rectangle ++(1,1);
     }

    \foreach \x in {0,...,4}{
      \fill[green] (\x,19) rectangle ++(1,1);
     }

    % --- Threshold line (12th-row edge) + label (ONLY here) ---
    \draw[red, line width=2pt] (0,12) -- (20,12);
    \node[anchor=east, red, font=\bfseries\small, xshift=-4pt] at (0,12) {X};
  \end{scope}

  %======================
  % PANEL 2 (shifted right by 22 units → 2-unit gap)
  %======================
  \begin{scope}[shift={(22,0)}]
    \draw[gray!60] (0,0) grid (20,20);
    \draw[thick] (0,0) rectangle (20,20);

    \foreach \x in {0,...,19}{
      \foreach \y in {0,...,9}{
        \fill (\x,\y) rectangle ++(1,1);
      }
    }

    \foreach \x in {0,...,11}{
      \fill (\x,10) rectangle ++(1,1);
     }

    \foreach \x in {0,...,4}{
      \fill[green] (\x,11) rectangle ++(1,1);
     }

    \foreach \x in {0,...,7}{
      \fill[orange] (\x,19) rectangle ++(1,1);
     }

    % --- Threshold line (12th-row edge) ---
    \draw[red, line width=2pt] (0,12) -- (20,12);
  \end{scope}

  %======================
  % PANEL 3 (shifted right again by 22 units)
  %======================
  \begin{scope}[shift={(44,0)}]
    \draw[gray!60] (0,0) grid (20,20);
    \draw[thick] (0,0) rectangle (20,20);

    \foreach \x in {0,...,19}{
      \foreach \y in {0,...,9}{
        \fill (\x,\y) rectangle ++(1,1);
      }
    }

    \foreach \x in {0,...,11}{
      \fill (\x,10) rectangle ++(1,1);
     }

    \foreach \x in {0,...,4}{
      \fill[green] (\x,11) rectangle ++(1,1);
     }

    \foreach \x in {12,...,19}{
      \fill[orange] (\x,10) rectangle ++(1,1);
     }

    % --- Threshold line (12th-row edge) ---
    \draw[red, line width=2pt] (0,12) -- (20,12);
  \end{scope}

\end{tikzpicture}%
}
\caption{X-threshold heuristic}\label{fig:X-heuristic}
\end{figure}

Under the X-threshold heuristic, a single value \(X \in \{0,\dots,p^{cap}\}\) is chosen to limit how many EVs may begin charging immediately upon arrival. The procedure moves forward in time, examining one slot \(j\) at a time. When new EVs arrive at slot \(j\), they are considered in their arrival order and assigned to start at \(j\) until the column reaches load \(X\). Any remaining EVs with \(ar_i = j\) are then scheduled using Row-filling: for each one, the earliest feasible start time \(\ge j\) within \([ar_i, d_i]\) that accommodates its entire \(l_i\)-block under capacity is selected. Once all arrivals at \(j\) are placed, the method proceeds to \(j{+}1\), leaving previous assignments unchanged. The heuristic does not dictate how \(X\) should be chosen; any reasonable rule can be adopted. Figure~\ref{fig:X-heuristic} illustrates the approach. We study two choices for \(X\):

\begin{itemize}
    \item \(\alpha\)-threshold: let \(\bar p\) be the mean number of EVs charging per time slot computed from historical logs or from simulated data on comparable days; set $X_{\alpha} \;=\; \lceil \bar p \rceil$.
    \item \(\beta\)-threshold (mean oracle peak): solve the oracle (perfect-information) model on each historical or simulated instance \(d\in\mathcal{D}\) and let \(p^{\max}_{\mathrm{oracle}}(d)=\max_{j\in\mathcal{J}} p^{\mathrm{oracle}}_j(d)\); set $X_{\beta}\;=\;\left\lceil \frac{1}{|\mathcal{D}|}\sum_{d\in\mathcal{D}} p_{\mathrm{oracle}}^{\max}(d) \right\rceil$.
\end{itemize}

\subsection{Learning-Based Approaches}

Learning-based approaches aim to derive adaptive policies that emulate expert scheduling behavior and generalize to unseen arrival patterns. Two complementary strategies are investigated: a standard \textit{SL-based approach} and its enhanced variant, \textit{DAgger}.

\subsubsection{Training Data and Expert Demonstrations}

Expert demonstrations are generated by solving the static optimization model \eqref{model_1} under multiple EV arrival scenarios. Each solved instance yields a sequence of \textit{state-action} pairs $(s_i, a_i)$ that describe the process of placing a charging block at its optimal position according to the perfect-information solution.

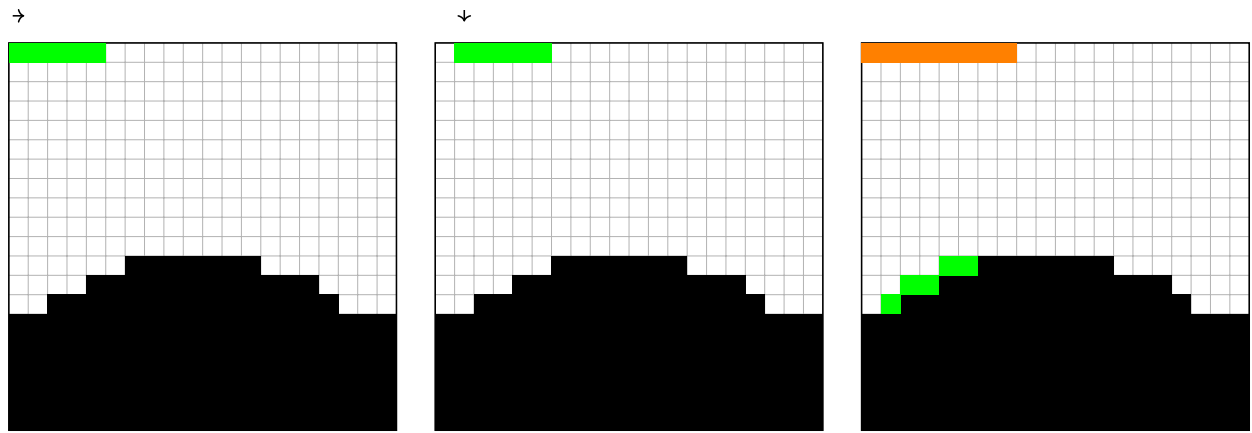
\begin{figure}[H]
\centering
\resizebox{\textwidth}{!}{%
\begin{tikzpicture}[scale=0.35]

  %======================
  % PANEL 1
  %======================
  \begin{scope}[shift={(0,0)}]
    \draw[gray!60] (0,0) grid (20,20);
    \draw[thick] (0,0) rectangle (20,20);

    % Half fill: bottom 10 rows (y=0..9), all columns (x=0..19)
    \foreach \x in {0,...,19}{
      \foreach \y in {0,...,5}{
        \fill (\x,\y) rectangle ++(1,1);
      }
    }

    \foreach \x in {2,...,16}{
      \foreach \y in {6,...,6}{
        \fill (\x,\y) rectangle ++(1,1);
      }
    }

    \foreach \x in {4,...,15}{
      \foreach \y in {7,...,7}{
        \fill (\x,\y) rectangle ++(1,1);
      }
    }

    \foreach \x in {6,...,12}{
      \foreach \y in {8,...,8}{
        \fill (\x,\y) rectangle ++(1,1);
      }
    }

%    \foreach \x in {0,...,11}{
%      \fill (\x,10) rectangle ++(1,1);
%     }
     %\end{tikzpicture}

    \foreach \x in {0,...,4}{
      \fill[green] (\x,19) rectangle ++(1,1);
     }

    % --- Trajectory arrows (Panel 1 only): right from x=0..12, then down at x=13 ---
%    \foreach \x in {0,...,12}{
%      \draw[->, line width=0.8pt] (\x+0.2,19.5) -- (\x+0.8,19.5);
%    }
%    \draw[->, line width=0.8pt] (13.5,19.8) -- (13.5,19.2);

    % --- action lane outside the image (Panel 1) ---
%    \draw[thick] (-0.5,21) -- (20.5,21);   % separator above the grid
    \draw[->, line width=0.8pt] (0.2,21.4) -- (0.8,21.4);   % one step to the right

    % --- EXTRA CELLS (Panel 1): add more black squares here if needed ---
    % \fill (2,12) rectangle ++(1,1);
    % \fill (7,15) rectangle ++(1,1);
  \end{scope}

  %======================
  % PANEL 2 (shifted right by 22 units → 2-unit gap)
  %======================

  \begin{scope}[shift={(22,0)}]
    \draw[gray!60] (0,0) grid (20,20);
    \draw[thick] (0,0) rectangle (20,20);

    % Half fill: bottom 10 rows (y=0..9), all columns (x=0..19)
    \foreach \x in {0,...,19}{
      \foreach \y in {0,...,5}{
        \fill (\x,\y) rectangle ++(1,1);
      }
    }

    \foreach \x in {2,...,16}{
      \foreach \y in {6,...,6}{
        \fill (\x,\y) rectangle ++(1,1);
      }
    }

    \foreach \x in {4,...,15}{
      \foreach \y in {7,...,7}{
        \fill (\x,\y) rectangle ++(1,1);
      }
    }

    \foreach \x in {6,...,12}{
      \foreach \y in {8,...,8}{
        \fill (\x,\y) rectangle ++(1,1);
      }
    }

%    \foreach \x in {0,...,11}{
%      \fill (\x,10) rectangle ++(1,1);
%     }
     %\end{tikzpicture}

    \foreach \x in {1,...,5}{
      \fill[green] (\x,19) rectangle ++(1,1);
     }

    % --- action lane outside the image (Panel 2) ---
%    \draw[thick] (-0.5,21) -- (20.5,21);            % separator above the grid
    \draw[->, line width=0.8pt] (1.5,21.7) -- (1.5,21.1);   % one step down

    % --- Trajectory arrows (Panel 1 only): right from x=0..12, then down at x=13 ---
%    \foreach \x in {0,...,12}{
%      \draw[->, line width=0.8pt] (\x+0.2,19.5) -- (\x+0.8,19.5);
%    }
%    \draw[->, line width=0.8pt] (13.5,19.8) -- (13.5,19.2);

    % --- EXTRA CELLS (Panel 1): add more black squares here if needed ---
    % \fill (2,12) rectangle ++(1,1);
    % \fill (7,15) rectangle ++(1,1);
  \end{scope}

  %======================
  % PANEL 3 (shifted right by 22 units → 2-unit gap)
  %======================
  \begin{scope}[shift={(44,0)}]
    \draw[gray!60] (0,0) grid (20,20);
    \draw[thick] (0,0) rectangle (20,20);

    \foreach \x in {0,...,19}{
      \foreach \y in {0,...,5}{
        \fill (\x,\y) rectangle ++(1,1);
      }
    }

    \foreach \x in {2,...,16}{
      \foreach \y in {6,...,6}{
        \fill (\x,\y) rectangle ++(1,1);
      }
    }

    \foreach \x in {4,...,15}{
      \foreach \y in {7,...,7}{
        \fill (\x,\y) rectangle ++(1,1);
      }
    }

    \foreach \x in {6,...,12}{
      \foreach \y in {8,...,8}{
        \fill (\x,\y) rectangle ++(1,1);
      }
    }

    \foreach \x in {1,...,1}{
      \foreach \y in {6,...,6}{
        \fill (\x,\y)[green] rectangle ++(1,1);
      }
    }

    \foreach \x in {2,...,3}{
      \foreach \y in {7,...,7}{
        \fill (\x,\y)[green] rectangle ++(1,1);
      }
    }

    \foreach \x in {4,...,5}{
      \foreach \y in {8,...,8}{
        \fill (\x,\y)[green] rectangle ++(1,1);
      }
    }

    \foreach \x in {0,...,7}{
      \fill[orange] (\x,19) rectangle ++(1,1);
     }

%    \foreach \x in {0,...,7}{
%      \fill[orange] (\x,19) rectangle ++(1,1);
%     }    

    % --- EXTRA CELLS (Panel 2) ---
    % \fill (3,12) rectangle ++(1,1);
    % \fill (10,13) rectangle ++(1,1);
  \end{scope}

\end{tikzpicture}%
}
%\caption{Row-Filling heuristic.}
\caption{Optimal Trajectory}\label{fig:agent-trajectory}
\end{figure}

In the gamified environment, the \textit{state} $s_i$ corresponds to an \textit{image} of the current grid, representing both the existing load distribution and the feasible region of the arriving EV. The \textit{action} $a_i$ corresponds to a \textit{joystick movement}--\textit{left}, \textit{right}, or \textit{down}--that incrementally adjusts the block's position. A full trajectory of consecutive state-action pairs therefore reproduces the sequence of movements leading to the block's optimal placement. These expert-generated trajectories form the labeled dataset used to train learning models that imitate expert decisions and generalize to unseen arrival patterns. Figure \ref{fig:agent-trajectory} illustrates the principle. The left frame shows the state of the system at the start of epoch $i$, with the arriving EV highlighted in green and the action ``move right'' shown above the grid. The middle frame shows the state after the block has moved to the right; at this point it must go down, and the arrow above the frame shows that downward action. The right frame shows the final state, after the block has been placed in its optimal position.

\subsubsection{Supervised Learning (SL)}\label{CNN}

In the SL framework, a neural network $f_\theta$ is trained to predict the expert action $a_i$ from each observed state $s_i$. Training minimizes the cross-entropy loss between predicted and demonstrated actions. Once trained, the model acts as a policy that, upon each EV arrival, observes the current grid and outputs the next movement command. This direct imitation of expert behavior allows near-instantaneous decision-making without explicit optimization.

However, because the model learns only from expert trajectories, it remains confined to the distribution of states visited by the expert. When executed online, it may encounter unobserved states, leading to error accumulation over time. This \textit{distributional drift} motivates the adoption of DAgger, which progressively exposes the model to its own induced states and corrects its behavior through additional expert feedback.

\subsubsection{Dataset Aggregation (DAgger)}\label{dagger}

The DAgger algorithm extends supervised learning by iteratively aligning the training and deployment distributions \citep{ross2011dagger}. After an initial SL phase, the trained policy is rolled out in simulation to generate new trajectories. Whenever the model visits a state where its decision diverges from the expert, the corresponding expert action is queried and appended to the dataset. The model is then retrained on this expanded corpus, blending expert and self-generated experience. Repeating this process over several iterations yields a robust policy that performs reliably under states unseen during pure imitation.

Within the gamified OCCSP, this iterative refinement is particularly beneficial: because block placement is sequential, small deviations early in a trajectory can propagate into large imbalances later on. DAgger mitigates this effect by reinforcing correct decisions in those previously unseen situations, improving both stability and generalization.

In each outer iteration, the agent generates an \emph{episode}, which consists of a full day or $N$ EVs, in the game environment under its current policy, yielding a sequence of state-action pairs. Let $s_i$ denote the state for decision epoch $i$ and let $I_i := \psi(s_i)$ be the RGB image of the game environment. Let $\mathbf{x}=(x_{ij})$ and $\mathbf{z}=(z_{ij})$ be the binary scheduling variables of Model \eqref{model_1} with feasible region $\mathcal{X}$. Holding all previously scheduled decisions fixed as in constraints \eqref{eq:fixed-past}, we complete the remaining portion of the episode to optimality by solving
\begin{equation}\label{eq:oracle-completion}
(\mathbf{x}_i^\star,\mathbf{z}_i^\star)\in
\arg\min_{(\mathbf{x},\mathbf{z})\in\mathcal{X}_i} F(\mathbf{x},\mathbf{z}),
\end{equation}
where $\mathcal{X}_i$ is defined as:
\begin{equation}\label{eq:Xi}
\mathcal{X}_i := \Bigl\{\, (\mathbf{x},\mathbf{z})\in\mathcal{X} \;:\;
x_{\iota j}=\bar{x}_{\iota j},\ \ z_{\iota j}=\bar{z}_{\iota j}
\quad \forall\, \iota\in\Ilt,\ \forall\, j\in\Jset \Bigr\}.
\end{equation}

Here, $F(\mathbf{x},\mathbf{z})$ is the load-balancing objective from Model \eqref{model_1}, \ie the objective that minimizes the spread $p^{\max}-p^{\min}$ over the horizon. We evaluate this objective on a state $s_i$ by fixing all past decisions and optimizing over the remaining EVs. At this point we assume perfect (oracle) knowledge of the remaining arrivals and their time windows, so the completion problem can be solved to optimality from $s_i$.

The expert label for the current block is the first control implied by this optimal completion, $a_i^{\mathrm{exp}}=\Phi\!\big(s_i,\mathbf{x}_i^\star,\mathbf{z}_i^\star\big)\in\{0,1,2\}$, representing the three possible actions (move left, move right, and move down). These labels are then made \emph{consistent with} the observed gameplay trajectory to produce corrected state-action pairs: each input tensor is the RGB image $I_i$ immediately after a left/right/down move, and each action is the corresponding expert movement extracted from $(\mathbf{x}_i^\star,\mathbf{z}_i^\star)$. These corrected demonstrations are aggregated \emph{episode-wise}: corrections are computed at every decision epoch, but appended to the training dataset after the $\eta$ episodes finish; in the main loop, we then retrain after adding the $\eta$ newly corrected episodes.

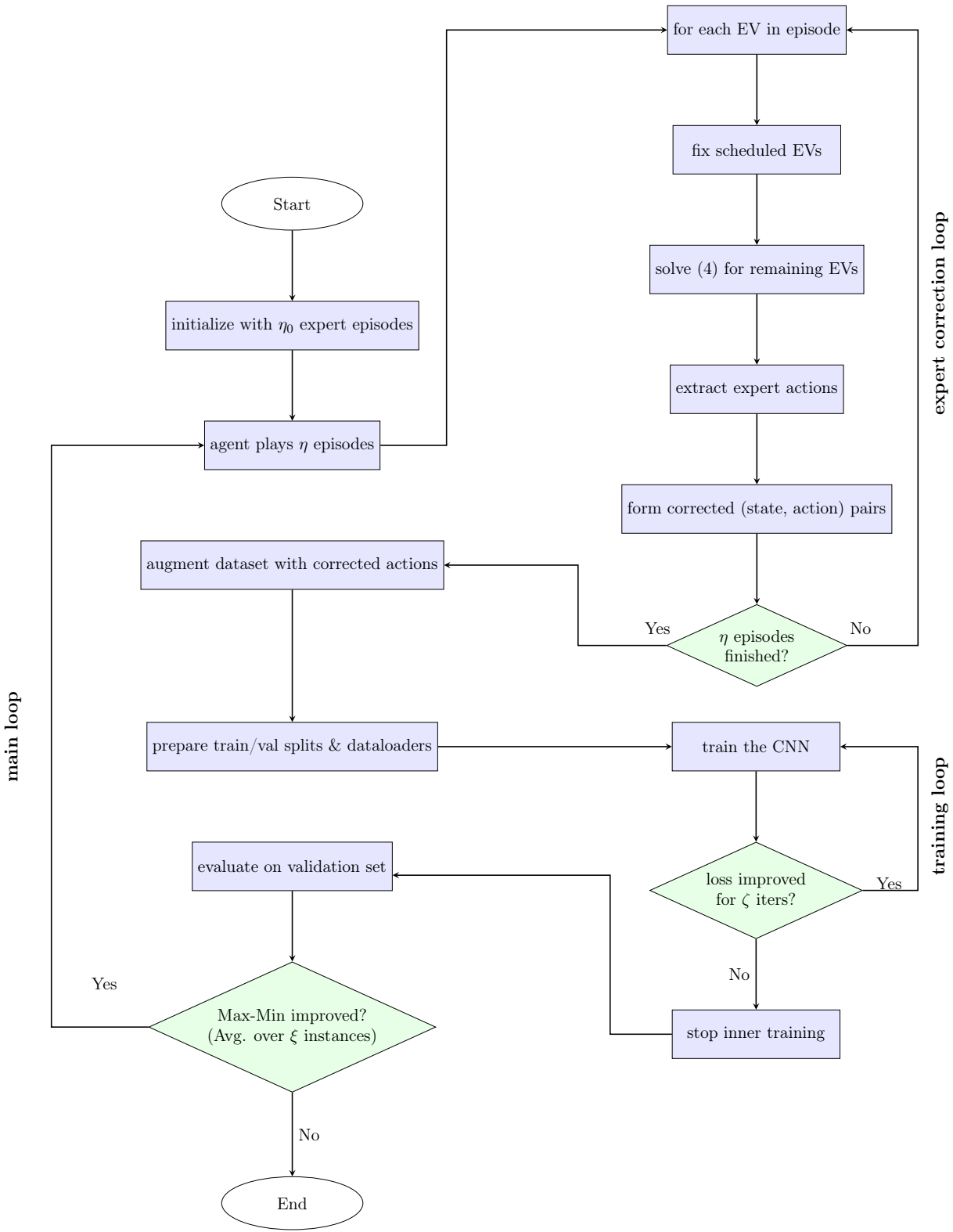
\begin{figure}[H]
\centering
\scalebox{0.75}{
\begin{tikzpicture}[node distance=1.6cm and 2.5cm]
% --- Add a terminal style for Start/End ---
\tikzset{terminal/.style={draw, ellipse, minimum width=3.2cm, minimum height=1.2cm, align=center}}

% Main vertical flow
\node (start0) [terminal] {Start};
\node (start) [process, below=of start0] {initialize with $\eta_{0}$ expert episodes};
\node (agentplay) [process, below=of start] {agent plays $\eta$ episodes};
\node (aggregation) [process, below=of agentplay] {augment dataset with corrected actions};
%\node (training) [process, below=of aggregation, minimum height=2.5cm] {supervised training on aggregated data};
%\node (training) [process, below=of aggregation] {supervised training on aggregated data};
%\node (training) [process, below=2.6cm of aggregation] {supervised training on aggregated data};
\node (training) [process, below=3cm of aggregation] {prepare train/val splits \& dataloaders};
\node (evaluation) [process, below=of training] {evaluate on validation set};
\node (improve) [decision, below=of evaluation] 
  {\shortstack{Max-Min improved?\\(Avg. over $\xi$ instances)}};
\node (stop) [terminal, below=of improve, yshift=-0.3cm] {End};

% Main arrows
\draw [arrow] (start0) -- (start);
\draw [arrow] (start) -- (agentplay);
\draw [arrow] (aggregation) -- (training);
%\draw [arrow] (aggregation.south) -- ++(0,-2.8) -- (training.north);
%\draw [arrow] (training) -- (evaluation);
\draw [arrow] (evaluation) -- (improve);
\draw [arrow] (improve) -- node[right] {No} (stop);
\draw [arrow] (improve.west) -- ++(-2.2,0.0) |- node[pos=0.5, xshift=12mm, yshift=-122mm] {Yes} (agentplay.west);

% Inner training loop (mirrored to the RIGHT of training box)
%\node (epoch) [process, right=2.8cm of training] {train the CNN};
\node (epoch) [process, right=5.3cm of training] {train the CNN};
\node (valcheck) [decision, below=of epoch] 
  %{\shortstack{cross-entropy loss improved?}};
  {\shortstack{loss improved\\for $\zeta$ iters?}};
\node (stopinner) [process, below=of valcheck] {stop inner training};

\draw [arrow] (training.east) -- (epoch.west);
\draw [arrow] (epoch) -- (valcheck);
\draw [arrow] (valcheck) -- node[left] {No} (stopinner);
%\draw [arrow] (stopinner.west) -- ++(-1.4,0) |- ([yshift=-0.2cm]training.east);
\draw [arrow] (stopinner.west) -- ++(-1.4,0) |- ([yshift=-0.2cm]evaluation.east);

\draw [arrow] (valcheck.east) -- ++(1.25,0) |- node[pos=0.5, xshift=-6.5mm, yshift=-31mm] {Yes} (epoch.east);

%\draw [arrow] (evcheck.east) -- ++(1.6,0) node[pos=0.5, xshift=-5mm, yshift=4mm] {No} |- (foreach.east);

% Expert correction loop (mirrored to the UPPER FAR RIGHT)
\node (foreach) [process, above right=8.3cm and 6.5cm of agentplay] {for each EV in episode};
\node (fixevs) [process, below=of foreach] {fix scheduled EVs};
\node (solveopt) [process, below=of fixevs] {solve \eqref{eq:oracle-completion} for remaining EVs};
\node (genact) [process, below=of solveopt] {extract expert actions};
\node (align) [process, below=of genact] {form corrected (state, action) pairs};
\node (evcheck) [decision, below=of align] {\shortstack{$\eta$ episodes\\finished?}};

% Entry to expert correction (mirrored)
\draw [arrow] (agentplay.east) -- ++(1.5,0) |- (foreach.west);

% Correction steps
\draw [arrow] (foreach) -- (fixevs);
\draw [arrow] (fixevs) -- (solveopt);
\draw [arrow] (solveopt) -- (genact);
\draw [arrow] (genact) -- (align);
\draw [arrow] (align) -- (evcheck);
\draw [arrow] (evcheck.east) -- ++(1.6,0) node[pos=0.5, xshift=-5mm, yshift=4mm] {No} |- (foreach.east);
\draw [arrow] (evcheck.west) -- ++(-2.0,0) node[pos=0.30, xshift=4mm, yshift=4mm] {Yes} |- (aggregation.east);

% Loop labels (mirrored positions)
\node[rotate=90, anchor=south, font=\bfseries\large] at ([xshift=-8cm, yshift=-6.6cm]agentplay.east) {main loop};
\node[rotate=90, anchor=south, font=\bfseries\large] at ([xshift=13cm, yshift=3cm]agentplay.east) {expert correction loop};
%\node[rotate=90, anchor=south, font=\bfseries\large] at ([xshift=11cm, yshift=-8cm]agentplay.east) {training loop};

\node[rotate=90, anchor=south, font=\bfseries\large] at ([xshift=13cm, yshift=-8.4cm]agentplay.east) {training loop};

\end{tikzpicture}
}
\caption{Overview of the DAgger training process}
\label{fig:dagger_diagram}
\end{figure}

Figure \ref{fig:dagger_diagram} illustrates this process. The flow diagram begins with the collection of expert demonstrations that serve as the initial training data based on Model \eqref{model_1}. An SL policy is first trained on this dataset and then deployed in the simulated environment, where it generates additional trajectories. Expert feedback is then obtained for each EV block by solving Model \eqref{eq:oracle-completion}, and the newly labeled pairs are added to the dataset. The model is retrained on the aggregated data, producing a refined policy that repeats the cycle until convergence. This loop, shown in the diagram, captures the essence of the DAgger procedure: continuous interaction between the learner and the expert to gradually close the gap between imitation and deployment performance.

\section{Experimental Results}\label{results}

To evaluate the proposed EV scheduling approaches, we design a comprehensive set of experiments organized into four main parts. Each part targets a distinct research question, focusing respectively on the value of gamification, the performance of various scheduling approaches under realistic conditions, the robustness of these approaches to data variability, and the economic advantage of the proposed approaches.

To ensure realism, the experiments are based on real-world data from the GMA. We first approximate the regional EV inventory and the number of local distribution feeders. From these estimates, we derive the implied average number of EV charging sessions per feeder, which serves as the experimental workload in our scenarios.

In the GMA, the estimated EV inventory is about 172,000 vehicles, obtained by summing the most recent regional counts for the Montréal urban agglomeration (\(\approx 44{,}000\) EVs), Laval (\(\approx 15{,}000\)), Montérégie (\(\approx 62{,}700\)), Lanaudière (\(\approx 23{,}900\)), and Laurentides (\(\approx 26{,}000\)) \citep{gouvqc_vze_2025}. Consistent with Hydro-Québec’s report that 80–90\% of EV charging in Québec occurs at home or at work \citep{hydroqc_ev_charging}, we assume a home/work charging (\ie not at public stations) share of 85\% in our analysis.

To translate vehicle counts into grid demand, we assume standard Hydro-Québec 25 kV three-phase distribution feeders, the typical configuration in Québec’s distribution network \citep{hydroqc_conditions_25kv_2021}. We begin by estimating how many such feeders serve the Greater Montréal Area (GMA). Hydro-Québec’s 2024 substation inventory lists 16 substations on the island of Montréal and about 11 in Laval and on the South Shore \citep{hq_substation_inventory_2024}. Using conservative averages of 33 feeders per on-island substation %(from project data: Bélanger, 29 initial circuits; Hochelaga, 35 at launch)
and 25 per off-island substation %(reflecting smaller initial builds at several 315/25 kV projects;
\citep{hq_poste_belanger_2010,hq_poste_hochelaga_2025}, this yields a total of $16 \times 33 + 11 \times 25 = 803$ feeders. Not all of these substations are currently in service: Côte-Saint-Luc and Rockfield, two of the sixteen island sites, are still awaiting energization \citep{csl_power_upgrade_2025,hq_saraguay_rockfield_eis_2025}. Applying an in-service factor of $1 - 2/16 = 0.875$ gives $803 \times 0.875 \approx 703$ feeders, which we round to approximately 700 active 25 kV feeders across the GMA.

Combining the 85\% home/work charging share with the estimated 700 feeders in the GMA yields $\frac{172{,}000 \times 0.85}{700} \approx 208.86$ charging sessions per feeder per day. For scenario design, we round this value to 200, meaning that a feeder can reasonably accommodate on the order of 200 Level-2 charging sessions (7 kW, 240 V, 30–32 A) per day \citep{chargehub_power_2025} without exceeding the utility’s planning envelope.

All experiments are conducted over a 24-hour planning horizon starting at noon, discretized into 96 time slots of 15 minutes each. Each EV is represented as a charging block defined by its arrival time, departure time, and required charging duration, which must be scheduled within its feasible window (\ie between its arrival and departure times). Different arrival-time distributions are tested (see Table~\ref{tab:exp-design}), while departure times are drawn from a common distribution, $\mathcal{N}(72,6)$, centered at 6:00~AM the following day (\ie time slot 72) and truncated so they neither precede arrivals nor exceed the planning horizon. Charging durations are uniformly sampled between 1 and 22 time slots (\ie 15 minutes to 5.5 hours) and adjusted, if necessary, to fit entirely within each EV’s availability window.

For example, in Scenario~1 (Table~\ref{tab:exp-design}), arrival times for 200 EVs are sampled from $\mathcal{N}(24,12)$, corresponding to the typical 6:00~PM rush hour (\ie time slot 24), while departure times are drawn from $\mathcal{N}(72,6)$ and truncated as needed. Charging durations are then uniformly sampled as described above and adjusted to fit within each sampled arrival–departure window. The same procedure applies to the other scenarios, using their respective EV counts and arrival-time distributions listed in Table~\ref{tab:exp-design}.

Performance is evaluated over 100 generated instances per scenario using two metrics.
First, the Max-Min load difference is computed for each instance, and the mean across all 100 instances is reported along with its 95\% confidence interval. This metric captures the load imbalance across time slots and directly reflects the optimization objective.
Second, the root mean squared error (RMSE) of the load profile is calculated for each instance, and the mean and 95\% confidence interval are similarly reported. Lower values of both the mean Max–Min load difference and the mean RMSE indicate better load balancing.

In particular, the Max-Min load difference captures the extremal load imbalance and directly reflects the traditional OCCSP objective, whereas RMSE quantifies how far the realized load profile deviates from the target (the average load in our case) across all time slots \cite{valogianni2025toward}. Because it penalizes larger deviations more strongly, a lower RMSE corresponds to a trajectory that keeps the system closer to its desired operating regime for most of the horizon, which from a technical, operational, and economic standpoint implies fewer pronounced peaks, lower network losses, more predictable use of capacity, and ultimately a reduced need for costly grid reinforcements.

Having described the data, we now turn to the model architectures and hyperparameters. The final hyperparameters were selected through a grid search over candidate architectures and training configurations. For the CNN, we compared one to five convolutional layers and found that three layers consistently yielded the best performance in terms of the mean Max-Min load balancing objective across 10 test instances. The selected CNN consists of three convolutional layers with $3\times3$ kernels, each followed by batch normalization, ReLU activation, and $2\times2$ max pooling. The resulting representation is then passed through two linear layers (256 and 128 units), with dropout rates of 0.4 and 0.3 applied after each layer, respectively, to mitigate overfitting. The final output layer produces logits for the three movement actions, transformed into probabilities via a softmax. We train with Adam (learning rate $0.001$) and weighted cross-entropy to address class imbalance, for up to 100 iterations, with early stopping if the evaluation loss does not improve for 5 consecutive iterations. The dataset is split 80/20 into training and evaluation sets, and class weights are computed dynamically based on the frequency of actions in the training data.

For models that are based on vector input, we used an MLP, which also requires hyperparameter tuning; in particular, we tuned model capacity and regularization jointly by sweeping hidden sizes $\{128,256,512\}$ over $2$-$4$ layers, dropout $p\in[0,0.55]$, and batch size $\{128,512,1024,2048\}$. Using a held-out validation split with early stopping, we observed that moving from $2\times128$ to $2\times256$ delivered clear gains, while adding a third or fourth layer or widening to $512$ yielded only marginal improvements and induced overfitting unless dropout was increased (which slowed learning). Dropout showed a sweet spot around $p=0.2$-$0.3$; higher values hurt convergence. For batching, a size of $512$ balanced gradient noise and generalization, $128$ was too noisy, whereas $1024$-$2048$ tended to over-smooth updates. For the final architecture, we fixed the model at two hidden layers of $256$ units each with dropout $p=0.2$ after each hidden layer and batch size $512$, which produced stable training and strong validation performance. The training setup (optimizer, loss, early stopping, and data split) mirrors the CNN configuration described above.

For the DAgger procedure, we varied the number of initial expert episodes ($\eta_{0}$) and agent rollout episodes per iteration ($\eta$), testing configurations of $(1,1)$, $(5,5)$, $(10,10)$, $(15,15)$, and $(20,20)$. The final setting of 10 initial expert episodes and 10 rollout episodes per iteration was selected based on its superior mean performance on the same evaluation metric. For the DAgger experiments, we use the same CNN/MLP architecture and training setup as in the supervised learning baseline. The updated agent is evaluated on $\xi=10$ held-out validation instances to assess whether the Max--Min load balancing objective improves.

\begin{table}[H]
\centering
\caption{Overview of the experimental scenarios.}
\label{tab:exp-design}
\begin{tabularx}{\textwidth}[t]{lccX}
\toprule
\textbf{Scenario} & \textbf{\# EVs} & \textbf{Arrival Distribution} & \textbf{Description} \\
\midrule
1 & 200 & $\mathcal{N}(24,12)$ & Typical evening rush hour, peak at 6:00~PM \\
2 & 300 & $\mathcal{N}(24,12)$ & Higher overall demand \\
3 & 200 & $\mathcal{N}(24,6)$  & Reduced arrival variability \\
4 & 200 &  \begin{tabular}[t]{c}
    50\% $\mathcal{N}(16,3)$ \\  
    50\% $\mathcal{N}(32,3)$
    \end{tabular}  & Bimodal demand with two peaks, 4:00~PM and 8:00~PM \\
\bottomrule
\end{tabularx}
\end{table}

\subsection{What Is the Value of Gamification?}

\begin{figure}[H]
    \centering
    \includegraphics[width=.75\textwidth]{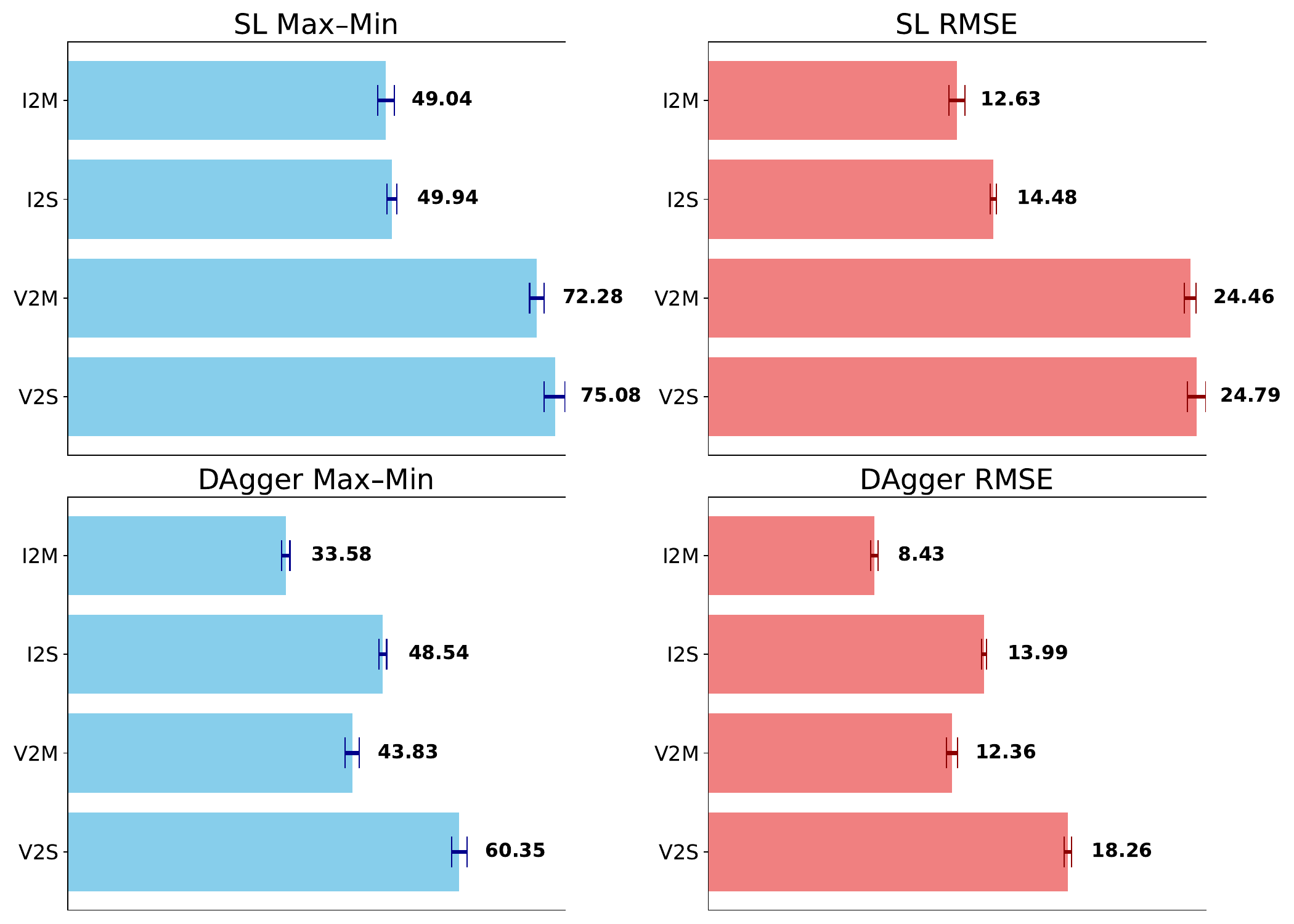}
    \caption{Comparing models' structures on Scenario 1}
    \label{fig:performance_comparison_2x2_1}
\end{figure}

We first assess the impact of gamified learning on scheduling performance by comparing the four model variants introduced in Section~\ref{gamification}, I2M, I2S, V2M, and V2S, under both SL and DAgger training based on Scenario 1.
%Performance is evaluated using two metrics: the Max-Min load difference, which quantifies load imbalance across time slots and directly reflects the optimization objective, and the Root Mean Squared Error (RMSE) of the load profile, which measures its overall deviation from the mean. Lower values in both metrics indicate better load balancing.
Figure \ref{fig:performance_comparison_2x2_1} reports the mean and 95\% confidence intervals of these metrics across the same 100 test instances.
For models sharing the same input representation, movement-based outputs yield clear and statistically significant gains over direct schedule-vector predictions. Under SL, both I2M and I2S achieve similar Max-Min values, but I2M exhibits a notably lower RMSE, showing that sequential movement decisions produce smoother load profiles. The same pattern holds for vector input: V2M consistently outperforms V2S on the Max-Min metric, with comparable RMSE values. These advantages are amplified under DAgger, where both I2M and V2M achieve lower means and non-overlapping confidence intervals relative to their schedule-vector counterparts. This confirms the benefit of framing the scheduling task as a sequence of discrete movements rather than a one-step classification problem.

Holding the output structure fixed, image-based inputs provide substantial and statistically significant improvements over vector-based representations. Under SL, I2M achieves much lower Max-Min and RMSE values than V2M, with no overlap in confidence intervals, and the same holds for I2S relative to V2S. The superiority of image-based models persists under DAgger: both I2M and I2S maintain lower Max-Min and RMSE values than their vector-input analogs, demonstrating that the visual encoding of the system state captures spatial dependencies that are lost in vector form.

Across all configurations, DAgger improves performance over standard SL, narrowing variability and reinforcing the robustness of learned policies. Among the four formulations, the gamified I2M model achieves the best overall results in both metrics and training schemes. These findings validate the theoretical results of Section~\ref{valueofgame}, where the combination of image inputs and movement outputs was shown to offer the most favorable generalization guarantees. Empirically, the same combination yields superior load balancing and reduced variance, confirming the practical value of gamification for the OCCSP.

\subsection{How Does Gamification Compare to Alternative Solution Approaches?}\label{experiemnt2}

\begin{figure}[H]
    \centering
    \includegraphics[width=1\textwidth]{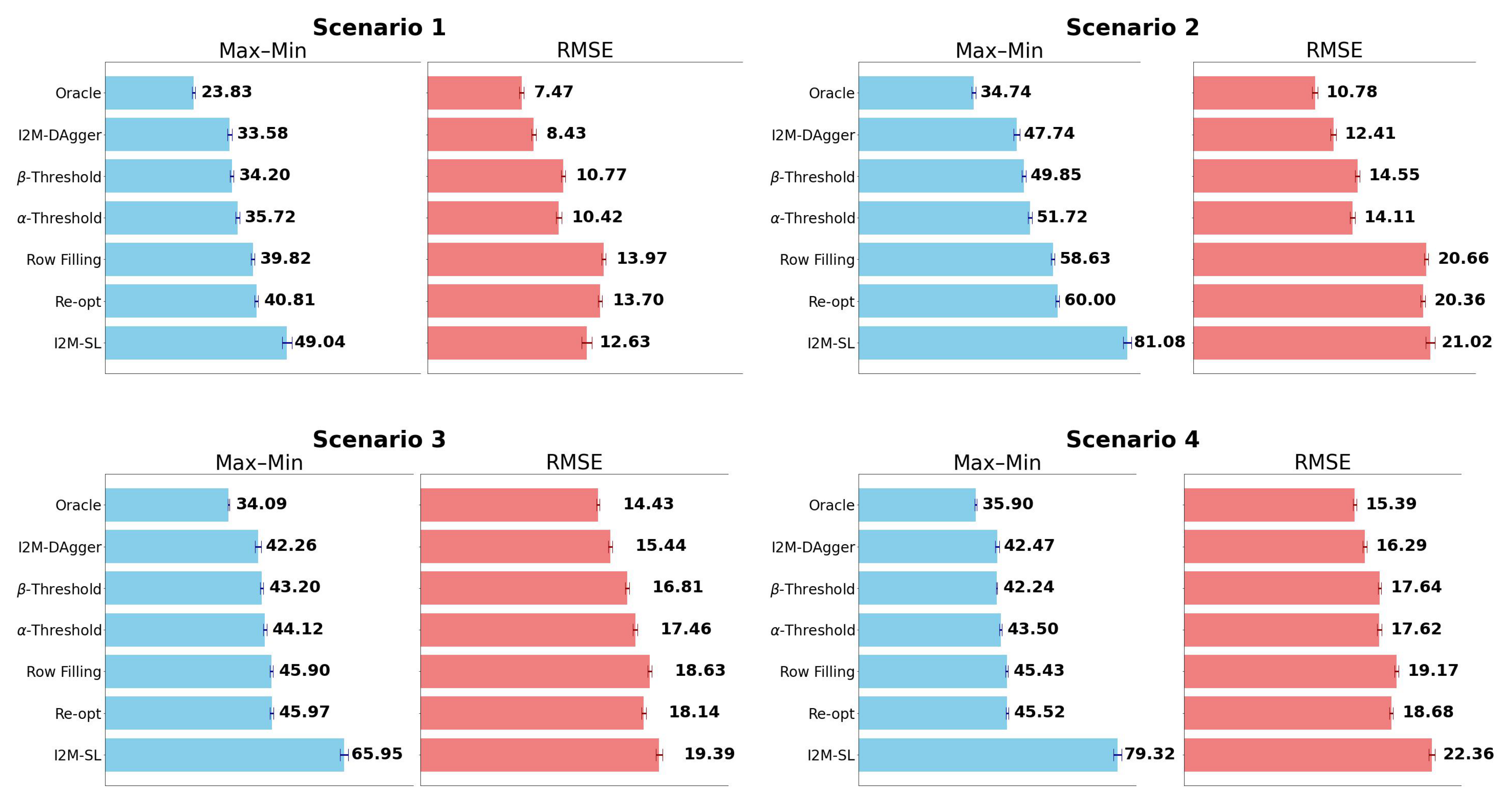}
    \caption{Comparisons across different arrival scenarios}
    \label{fig:performance_comparison_2x2_2}
\end{figure}

To evaluate the generalizability of the gamified I2M strategy beyond the 200-EV evening-rush-hour case, we benchmark five approaches, namely, the I2M-DAgger model, its supervised-learning counterpart (I2M-SL), the offline oracle, an online re-optimization (Re-opt) method, and three heuristic policies, across the four arrival scenarios summarized in Table~\ref{tab:exp-design}.
Figure~\ref{fig:performance_comparison_2x2_2} presents the mean and confidence intervals of each method across these scenarios. In every case, the offline oracle achieves the lowest Max-Min and RMSE values, providing a useful but unattainable benchmark that assumes perfect knowledge of all arrivals. The Re-opt baseline performs only marginally better than the simplest heuristics, as it optimizes each vehicle in isolation and does not anticipate future demand. The row-filling heuristic performs worst among heuristics, lacking any form of demand forecasting. The supervised I2M model also underperforms: trained solely on oracle trajectories, it cannot recover when early errors push the system into unfamiliar states. In contrast, the I2M-DAgger policy, which enriches its training data with expert corrections on suboptimal states, consistently outperforms both the $\alpha$- and $\beta$-threshold heuristics. In the 200-EV evening-rush scenario, its Max-Min lies clearly below both thresholds. Although the upper bound of the $\beta$-threshold’s confidence interval slightly overlaps the I2M-DAgger interval, the RMSE comparison shows no overlap, confirming a statistically significant reduction in load variance. Under heavier demand (300 EVs), I2M-DAgger dominates even more strongly: both its Max-Min and RMSE intervals lie entirely below those of the two threshold heuristics, indicating robust gains under higher congestion.
In the remaining cases, two-peak arrivals and reduced-variability demand, the Max-Min intervals of I2M-DAgger and the $\beta$-threshold heuristic overlap, suggesting similar peak-valley balancing. However, in both settings, the RMSE interval of I2M-DAgger remains entirely below that of $\beta$-threshold, confirming smoother load profiles despite comparable peak balancing performance.

Overall, these results show that across all four arrival patterns, the I2M-DAgger model consistently improves upon both learning and heuristic baselines, maintaining statistically significant advantages in load variance and overall balance. The findings demonstrate that imitation learning with dataset aggregation generalizes effectively beyond the nominal 200-EV scenario, preserving stable performance as demand intensity and temporal variability change.

\subsection{How Robust Is Gamification?}

We assess robustness by perturbing the parameters of Scenario 1 around their baseline values. Specifically, we take Scenario 1 (200 EVs, arrivals distributed as $\mathcal{N}(24,12)$) as the reference. For each sensitivity setting, we generate 100 instances as follows: for the parameter(s) under test (arrival mean, arrival variance, or number of EVs), we sample 100 perturbed values within a $\pm 10\%$ band around the baseline and construct the instances using these values, while keeping all other parameters fixed at their Scenario 1 levels. A fourth setting perturbs all three parameters simultaneously in the same manner. Both I2M-DAgger and I2M-SL are trained exclusively on the unperturbed Scenario 1 and evaluated on these perturbed instances.

\begin{figure}[H]
    \centering
    \includegraphics[width=1\textwidth]{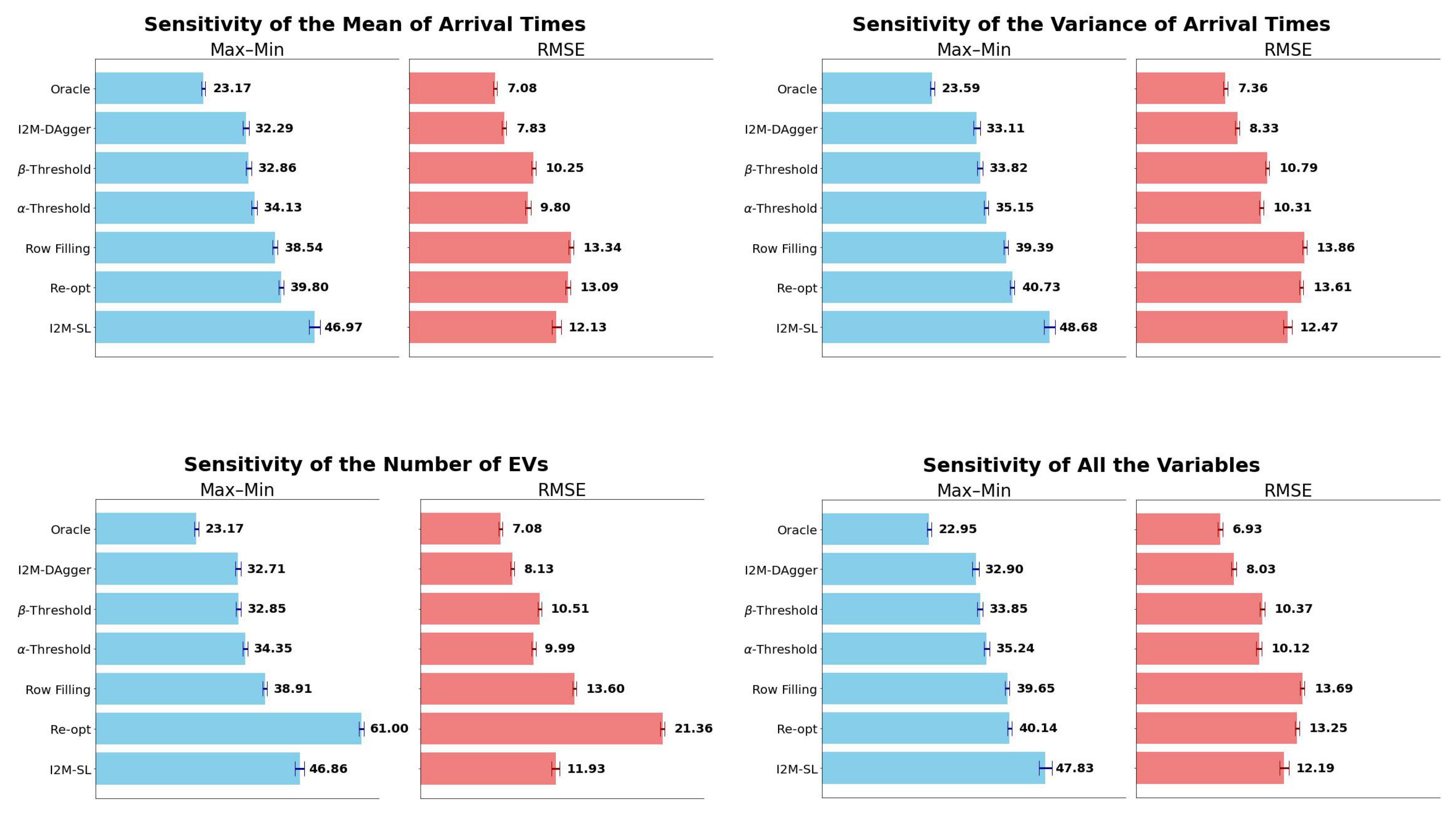}
    \caption{Sensitivity analyses}
    \label{fig:performance_comparison_2x2_3}
\end{figure}

The Results show that across all sensitivity settings, the oracle achieves the lowest Max-Min and RMSE and serves as a best-case reference, while I2M-SL and Re-opt define upper bounds on attainable performance and are used only to contextualize the gap with real-time strategies. The main comparison is therefore between I2M-DAgger and the heuristic policies. As shown in Figure~\ref {fig:performance_comparison_2x2_3}, I2M-DAgger achieves better or comparable Max-Min than the best heuristics, with confidence-interval overlap often observed with $\beta$-Threshold; this indicates similar peak-valley balancing. Crucially, in all scenarios, mean, variance, EV-count, and worst-case, I2M-DAgger outperforms the heuristics on RMSE with non-overlapping intervals, yielding smoother load profiles despite comparable Max-Min. In particular, under mean and variance perturbations, I2M-DAgger matches the best heuristic on Max-Min while delivering a statistically significant RMSE improvement; under EV-count and worst-case perturbations, the same pattern persists, indicating stable overall balance and adaptability. It is worth noting that, even though both I2M-DAgger and I2M-SL are trained only on unperturbed Scenario~1, they maintain strong performance on perturbed instances within a \( \pm 10\% \) band (a \(20\%\)-wide interval) around the baseline.

\subsection{What Are the Economic Benefits?}

In this section, we quantify the economic benefits of the proposed EV-scheduling methods by comparing their reductions in feeder peak demand with the prevailing practice, where each EV begins charging immediately upon plug-in called plug-in-to-charge \citep{martel2024smart}. We do so by expressing each sustained kilowatt of peak reduction in terms of Hydro-Québec’s avoided distribution capacity cost, as filed with the \textit{Régie de l’énergie}. This long-run metric, currently set at 164~CAD per kW-year (2025~\$), represents the annualized cost the utility would avoid if future upgrades to feeders, conductors, or distribution transformers can be deferred or downsized because of lower peak demand \citep{regie_R-4270-2024_B-0033,regie_R-4270-2024_A-0020}. Accordingly, a one-kilowatt reduction in the planning-year peak corresponds to an avoided cost of 164 CAD per year.

It is important to note that this avoided-capacity metric is an economic valuation tool rather than a direct prediction of near-term capital expenditure savings. The EVs in our scenarios already exist, and the distribution system already accommodates their uncoordinated charging behavior. Thus, coordinated charging does not retroactively eliminate investments that have already been made or committed. Instead, it represents a sustained peak reduction that, if occurring on feeders approaching or projected to reach their limits, could defer the timing or scale of future upgrades. The values reported below therefore quantify the \emph{potential} avoided-capacity benefit associated with lower EV-driven peaks, under standard planning assumptions, rather than guaranteed reductions in current utility spending.
With this interpretation in mind, we estimate the avoided-capacity value achieved by each scheduling method. In Scenario~1 (200 EVs per feeder over 24 hours), the Max Load is the average, over 100 instances, of the maximum number of EVs charging simultaneously. Under the prevailing immediate-charging practice, the Max Load is 75 EVs, whereas I2M-DAgger achieves 35 EVs, yielding a peak reduction of $75 - 35=40$ EVs. Assuming a per-EV charging power of $u = 7$~kW, this corresponds to a 280~kW reduction in peak demand per feeder. Valued at 164 CAD/kW-year, this represents a potential avoided capacity value of $280 \times 164 = 45{,}920$~CAD per feeder per year. Scaled across approximately 700 feeders in the GMA, this amounts to $45{,}920 \times 700 = 32{,}144{,}000$~CAD annually. The same calculation is applied to the remaining methods and scenarios.

\begin{figure}[H]
    \centering
    \includegraphics[width=.75\textwidth]{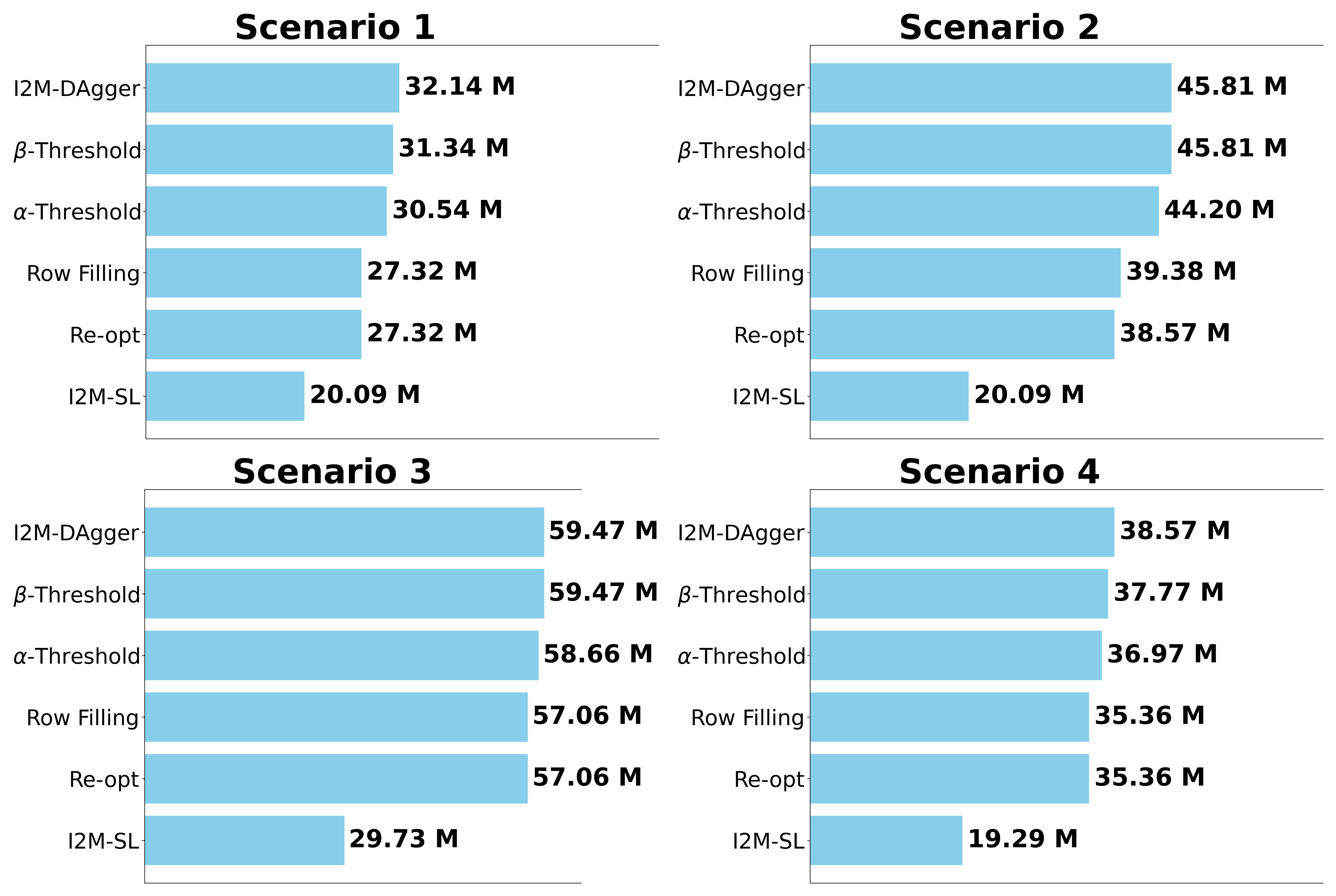}
    \caption{Cost Savings}
    \label{fig:economy_comparison_2x2_3}
\end{figure}

Figure~\ref{fig:economy_comparison_2x2_3} summarizes the results relative to the prevailing practice. Across all four scenarios, I2M-DAgger and $\beta$-Threshold consistently achieve the largest potential avoided-capacity value, followed closely by $\alpha$-Threshold, Row Filling, and Re-opt. I2M-SL performs noticeably worse. The ranking is consistent across scenarios, although the magnitude of the potential savings varies. Scenario~3 shows substantially higher values despite having the same daily EV count as Scenarios~1 and~4; its more concentrated arrival distribution leads uncoordinated charging to create significantly sharper peaks, which coordinated policies are able to reduce more effectively. This illustrates how the economic value of coordination depends not only on the total number of EVs but also on the temporal structure of demand.

\section{Conclusion}\label{conclusion}

The paper tackles online residential EV charging by proposing the OCCSP framework grounded in a game-inspired optimization approach. We cast the problem as an MDP that captures the sequential arrival of requests, their energy needs and deadlines, and capacity limits, and we implement this formulation as an interactive grid-based environment that visualizes the load-balancing task. 

We evaluate four model architectures, I2M, I2S, V2M, and V2S, that differ in input representation and output format. Theoretical analysis based on parameter growth and generalization bounds indicates that combining image inputs with sequential movement outputs (I2M) offers the most favorable sample-complexity and generalization properties. 

For learning, we use supervised training on oracle demonstrations and then apply DAgger to add expert corrections collected on off-policy states. The learned policies are compared with rule-based heuristics (Row-Filling, $\alpha$-Threshold, $\beta$-Threshold), a myopic online optimization baseline, and a full-information oracle. 

Across diverse arrival scenarios, the I2M policy trained with DAgger consistently outperforms the alternatives on both Max-Min load difference and RMSE, showing stronger generalization and better recovery from suboptimal states; these gains are validated via confidence-interval and instance-wise comparisons. 

From a managerial perspective, our results provide a concrete roadmap for utilities and aggregators navigating rapid EV adoption. In a proposed case study calibrated to Hydro-Québec’s cost parameters for the GMA, the best-performing methods (I2M-DAgger and $\beta$-Threshold) achieve comparable economic outcomes and clearly outperform the other baselines. For instance, relative to the status-quo “plug-in-to-charge” benchmark, the peak load drops from 75 to 35 EVs (a 53\% reduction), corresponding to a potential annual avoided cost of up to \$32.14 million under Scenario 1. The reduction is even more pronounced in Scenario 3, where the peak decreases from 116 to 42 EVs (a 63\% reduction), yielding up to \$59.47 million in annual savings. Overall, the results show that centralized orchestration of charging operations carried via gamified models is a valid alternative to balance load efficiently with modest computational effort, reducing pressure on costly grid upgrades and facilitating large-scale coordination. More broadly, the proposed framework generalizes to other online resource-allocation settings, offering decisions that are both robust to uncertainty and transparent to stakeholders.

%From a managerial perspective, the results point to a practical path for utilities and aggregators confronting rapid EV adoption: a gamified, expert-guided scheduler that is simple to operate and fast enough for real time. By shifting from price-based incentives \MC{Unsure about this first part...} or heavy optimization workflows to an interpretable, image-based interface, operators can balance load with modest computation, easing pressure on costly grid upgrades and enabling coordination at scale. More broadly, the same framework extends to other online resource-allocation problems, offering decisions that are both robust to uncertainty and transparent to stakeholders.

Beyond these results, our study adds evidence to a broader trend in hybrid analytics. Over the past decade, ML has been used both to solve combinatorial problems directly and to accelerate classical OR algorithms. Our findings confirm a complementary path: use strong OR solvers on static formulations as expert labelers to train ML policies for online variants. In practice, the optimizer supplies high-quality demonstrations; the learned policy internalizes that structure and executes instantaneously at run time. This OR-as-labeler, ML-for-online pattern is pragmatic, extensible, and, based on our results, well worth pursuing in other sequential decision problems.

\section{Acknowledgments}
This research was made possible through support from Calcul Québec (calculquebec.ca), the Digital Research Alliance of Canada (alliancecan.ca), the Institute for Data Valorization (IVADO), the Natural Sciences and Engineering Research Council of Canada (NSERC), and the Fonds de recherche du Québec – Nature et technologies (FRQNT).

%\bibliographystyle{plainnat}

%\bibliography{bibmain}

\bibliographystyle{informs2014}
\bibliography{bibmain}

\renewcommand{\thesection}{\Alph{section}}  % Section label: A, B, C...
\setcounter{section}{0}  

% Override how the section title is displayed:
\makeatletter
\renewcommand{\@seccntformat}[1]{Appendix~\csname the#1\endcsname\quad}
\makeatother

\clearpage
\section{Proofs}\label{sec:appendixA}

\subsection{Proof of Lemma \ref{lem:ev-param-gap}}

\cite{kohler2023analysis}, Lemma 13, derived a VC-dimension bound for CNN classes with ReLU activations, establishing the upper bound
\[
\kappa\,C^{2}\,\log(C\,\lambda),
\]
where $\kappa>0$ is a constant depending only on network-side parameters, and $C$ is the number of convolution layers. The term $\kappa\,C^{2}$ is constant with respect to $T$, with $\kappa$ independent of $T$, since global pooling ensures that the post-convolutional representation length does not grow with $T$. This yields the CNN-specific VC-dimension estimate
\[
\mathrm{VCdim}(\mathcal H_I)\;\le\;\kappa\,C^{2}\,\log(C\,\lambda)
\quad\Longrightarrow\quad
\Phi_I \;=\; \Theta\!\big(\log(\lambda)\big),
\]

On the other hand, for the vector-input MLP with input length $d=L+T$, $\eta$ hidden layers of width $m$, and output dimension $|\mathcal A|$, the total number of weights is
\[
W_V \;=\; d\,m + (\eta-1)m^2 + m\,|\mathcal A|
\;=\; mT + \underbrace{\big(mL+(\eta-1)m^2+m|\mathcal A|\big)}_{\text{independent of }T}.
\]

\cite{bartlett2019nearly} proved that the VC-dimension bound for MLPs is 
\[
\mathrm{VCdim}(\mathcal H_V) = \Theta(W\log W).
\]

Hence
\[
\Phi_V = W_V\log W_V \;=\;\Theta\!\big(T\log T\big)\quad(T\to\infty).
\]
%This dependence only comes from the $T$-dimensional per-slot part of the input; adding any other fixed-size features to the vector input does not change the $W_V = \Theta(T)$ and $W_V \log W_V = \Theta(T \log T)$ scaling.

We therefore have $\Phi_I=\Theta(\log(\lambda))$. If the image resolution $\lambda$ scales with the temporal discretization ($\lambda\asymp T$), then $\Phi_I=\Theta(\log T)$, while from above $\Phi_V= W_V\log W_V=\Theta(T\log T)$. Consequently,
\[
\Phi_I \;\ll\; \Phi_V
\quad\text{as }T\to\infty,
\]
which establishes the claim.

\hfill\(\square\)

\clearpage
\subsection{Proof of Theorem \ref{theory1}}

\paragraph{VC‐dimension bound} In Lemma~\ref{lem:ev-param-gap}, we established the VC‐dimension of two classes of models for vector and image inputs. Since we are comparing input modalities only, it is fair to assume that the output architectures are aligned; hence we apply the same generalization bound separately to each class.  

The {\em VC‐dimension} of a hypothesis class $\mathcal H$, denoted $\mathrm{VCdim}(\mathcal H)$, is the largest size of a finite set that can be {\em shattered} by $\mathcal H$. A standard uniform convergence result (see, \eg \cite{anthony2009neural}) states that for any class $\mathcal H$ of VC‐dimension $d$, with probability at least $1-\delta$ over $n$ samples,
\[
  R(h)\;\le\;\widehat R_n(h)\;+\;C\;\sqrt{\frac{\mathrm{VCdim}(\mathcal H) + \ln(\tfrac{1}{\delta})}{n}},
\]
where $C>0$ is a universal constant independent of $\mathcal H$, $n$, or $\delta$.  

\paragraph{(a) MLP with vector input}  
For a feedforward neural network, with architecture defined in Lemma \ref{lem:ev-param-gap}, with $W$ total parameters, it is known \citep{bartlett2019nearly} that
\[
  \mathrm{VCdim}(\mathcal H)\;=\;O\bigl(W\log W\bigr).
\]
Concretely, there exists $K>0$ such that
\[
  \mathrm{VCdim}(\mathcal H)\;\le\;K\,W\log W
  \quad\text{for all sufficiently large }W.
\]
Substituting $d=\mathrm{VCdim}(\mathcal H)\le K\,W\log W$ into the VC bound gives
\[
  R(h)\;\le\;\widehat R_n(h)\;+\;C\sqrt{\frac{K\,W\log W + \ln(\tfrac{1}{\delta})}{n}}.
\]
Factoring $K$ inside the square root and absorbing constants yields
\[
  R(h)\;\le\;\widehat R_n(h)\;+\;C'\,\sqrt{\frac{W\log W + \ln(\tfrac{1}{\delta})}{n}}.
\]
From Lemma~\ref{lem:ev-param-gap}, for vector input $W=\Theta(T)$ and hence $W\log W=\Theta(T\log T)$. Therefore,
\[
  R(h)\;\le\;\widehat R_n(h)\;+\;C'\,\sqrt{\frac{T\log T + \ln(\tfrac{1}{\delta})}{n}}.
\]

\paragraph{(b) CNN with image input}  
For the CNN class $\mathcal H_I$ with image input, Lemma~\ref{lem:ev-param-gap} gives
\[
  \mathrm{VCdim}(\mathcal H_I)\;\le\;\Phi_I\;=\;\Theta(\log(C\lambda)).
\]
Substituting $d=\mathrm{VCdim}(\mathcal H_I)\le \Phi_I$ into the VC bound yields
\[
  R(h)\;\le\;\widehat R_n(h)\;+\;C\,\sqrt{\frac{\Phi_I + \ln(\tfrac{1}{\delta})}{n}}.
\]
If the image resolution scales with $T$ (i.e., $\lambda \asymp T$), then $\Phi_I=\Theta(\log(C\lambda))=\Theta(\log T)$, so
\[
  R(h)\;\le\;\widehat R_n(h)\;+\;C''\,\sqrt{\frac{\log T + \ln(\tfrac{1}{\delta})}{n}},
\]
for some absolute constant $C''>0$.

\paragraph{Asymptotic comparison as $T\to\infty$}
Let
\[
E_{\mathrm{MLP}}(T)\;=\;C'\sqrt{\frac{T\log T+\ln(\tfrac{1}{\delta})}{n}},
\qquad
E_{\mathrm{CNN}}(T)\;=\;C''\sqrt{\frac{\log T+\ln(\tfrac{1}{\delta})}{n}}.
\]
With $n$ and $\delta$ fixed, 
\[
\frac{E_{\mathrm{CNN}}(T)}{E_{\mathrm{MLP}}(T)}
=\frac{C''}{C'}\sqrt{\frac{\log T+\ln(\tfrac{1}{\delta})}{T\log T+\ln(\tfrac{1}{\delta})}}
\;\xrightarrow[T\to\infty]{}\;\frac{C''}{C'}\cdot\frac{1}{\sqrt{T}}\;=\;0,
\]
where $E_{\mathrm{CNN}}(T)$ is the error bound contribution for CNNs and $E_{\mathrm{MLP}}(T)$ is the error bound contribution for MLPs. 

Hence, the CNN upper‐bound error term is asymptotically smaller than the MLP’s.

\hfill\(\square\)

\clearpage
\subsection{Proof of Theorem \ref{thm:MvT-complete}}

\paragraph{Master multiclass bound (Natarajan-dimension form)}  For any multiclass hypothesis class of Natarajan dimension \(d\), \(k\) output classes, and \(o\) i.i.d.\ training points, with probability at least \(1-\delta\),
\[
  R(h)\;\le\;\widehat R_o(h)\;+\;\sqrt{\frac{d\,(\ln o + \ln k)\;+\;B}{o}}.
\]
%(See, e.g., Kearns–Schapire 1994 or Bartlett \emph{et al.} 2019.)

\paragraph{Specialization to the two models}
\begin{itemize}
  \item \emph{Vector model (S classes).}  Here \(o=n\), \(k=S\), \(d=d_S\).  The bound yields
%    \[
%      R(h)\;\le\;\widehat R_n(h)\;+\;\sqrt{\frac{d_S\,(\ln n + \ln S)+B}{n}}
%      \quad\Longrightarrow\quad
%      \underbrace{\frac{d_S\,(\ln n + \ln S)+B}{n}}_{G_1}\,.
%    \]

    \[
    R(h)\;\le\;\widehat R_n(h)\;+\;\sqrt{\frac{d_S\,(\ln n + \ln S)+B}{n}},
\qquad
G_1 \;:=\; \frac{d_S\,(\ln n + \ln S)+B}{n}.
    \]
  \item \emph{Sequential model (M classes per step).}  Here \(o=nH\), \(k=M\), \(d=d_M\).  The bound yields
%    \[
%      R(h)\;\le\;\widehat R_{nH}(h)\;+\;\sqrt{\frac{d_M\,(\ln(nH) + \ln M)+B}{nH}}
%      \quad\Longrightarrow\quad
%      \underbrace{\frac{d_M\,(\ln n + \ln H + \ln M)+B}{nH}}_{G_2}\,.
%    \]

    \[
  R(h)\;\le\;\widehat R_{nH}(h)\;+\;\sqrt{\frac{d_M\,(\ln(nH) + \ln M)+B}{nH}},
  \qquad
  G_2 \;:=\; \frac{d_M\,(\ln n + \ln H + \ln M)+B}{nH}\,.   
    \]

\end{itemize}
The sequential model has the smaller leading “generalization term” precisely if
\[
  G_2 \;<\; G_1.
\]

By assuming $X=\ln n + \ln H$, we follow the following four steps:

\medskip\noindent\emph{Step 1: Form the ratio.}
\[
  \frac{G_2}{G_1}
  = \frac{\tfrac{d_M\,(X + \ln M) + B}{nH}}
         {\tfrac{d_S\,(\ln n + \ln S) + B}{n}}
  = \frac{d_M\,(X + \ln M) + B}{H\,\bigl(d_S\,(\ln n + \ln S) + B\bigr)}.
\]
Set
\[
  R_1 \;=\;\frac{d_M\,X + B}{\,d_S\,(\ln n + \ln S) + B\,},
  \quad
  R_2 \;=\;\frac{d_M\,\ln M}{\,d_S\,(\ln n + \ln S) + B\,}.
\]
Then
\[
  \frac{G_2}{G_1}
  = \frac{1}{H}\,\bigl(R_1 + R_2\bigr).
\]

\medskip\noindent\emph{Step 2: Bound \(R_1\).}  Write
\[
  D = d_S\,(\ln n + \ln S), 
  \quad
  E = B,
\]
so \(R_1=\frac{(d_M\,X+E)}{(D+E)}\).  Two cases:
\begin{itemize}
    \item If \(d_M \le d_S\), then \(d_M\,X \le d_S\,X\le D\) (since \(X=\ln n+\ln H\ge\ln n\)), so
  \[
    d_M\,X + E \;\le\; D + E
    \quad\Longrightarrow\quad
    R_1 \;\le\;1.
  \]
  \item If \(d_M > d_T\), define
  \(\alpha = \frac{D}{(D+E)}\in(0,1)\).  Then
  \[
    R_1
    = \frac{d_S\,X + E + (d_M-d_S)\,X}{D+E}
    = \alpha\cdot1\;+\;(1-\alpha)\;\frac{d_M}{d_S}
    \;\le\;\max\!\Bigl(1,\tfrac{d_M}{d_S}\Bigr).
  \]
\end{itemize}

Hence in all cases 
\[
  R_1 \;\le\;\max\!\Bigl(1,\,\tfrac{d_M}{d_S}\Bigr).
\]

\medskip\noindent\emph{Step 3: Bound \(R_2\).}
\[
  R_2
  = \frac{d_M\,\ln M}{D + E}
  \;\le\;
  \frac{d_M\,\ln M}{D}
  = \frac{d_M}{d_T}\;\frac{\ln M}{\ln n + \ln S}.
\]

\medskip\noindent\emph{Step 4: Combine.}
\[
  \frac{G_2}{G_1}
  \;\le\;
  \frac{1}{H}\Bigl[
    \max\!\bigl(1,\tfrac{d_M}{d_S}\bigr)
    \;+\;
    \frac{d_M}{d_S}\,\frac{\ln M}{\ln n + \ln S}
  \Bigr].
\]
Requiring \(\tfrac{G_2}{G_1}<1\) is equivalent to
\[
  \max\!\Bigl(1,\tfrac{d_M}{d_S}\Bigr)
  + \frac{d_M}{d_S}\,\frac{\ln M}{\ln n + \ln S}
  < H.
\]
Split into the two subcases:
\begin{itemize}
  \item If \(d_M\le d_S\), the condition becomes
\[
  r \;=\;\frac{d_M}{d_S}\;\le1
  \quad\Longrightarrow\quad
  r \,\frac{\ln M}{\ln N + \ln S}
  \;\le\;
  1 \,\frac{\ln M}{\ln N + \ln S}
  \;=\;
  \frac{\ln M}{\ln N + \ln S}.
\]
Hence
\[
  \max\!\bigl(1,r\bigr)
  + r\,\frac{\ln M}{\ln N + \ln S}
  \;\le\;
  1 \;+\;\frac{\ln M}{\ln N + \ln S}.
\]

Then
  
    \[
      1 + \frac{\ln M}{\ln n + \ln S} \;<\; H.
    \]

  \item If \(d_M> d_S\), it becomes
    \[
      \frac{d_M}{d_S}\,\Bigl(1 + \tfrac{\ln M}{\ln n + \ln S}\Bigr)
      < H
      \quad\Longleftrightarrow\quad
      \frac{d_M}{d_S}
      < \frac{H}{\,1 + \tfrac{\ln M}{\ln n + \ln S}\,}.
    \]
\end{itemize}

\[
r = \frac{d_M}{d_S},
\qquad
a = \frac{\ln M}{\ln N + \ln S},
\]
and we start from
\[
\max\bigl(1,r\bigr) + r\,a < H.
\]
Now split into two cases:

\[
\begin{aligned}
\text{Case 1: }r\ge1: 
&\quad \max(1,r)=r
\;\Longrightarrow\;
r + r\,a < H
\;\Longrightarrow\;
r\,(1+a)<H
\;\Longrightarrow\;
r<\frac{H}{1+a}.\\
\text{Case 2: }r\le1: 
&\quad \max(1,r)=1
\;\Longrightarrow\;
1 + r\,a < H
\;\Longrightarrow\;
1 + a < H
\;\Longrightarrow\;
r \le1 < \frac{H}{1+a}
\;\Longrightarrow\;
r<\frac{H}{1+a}.
\end{aligned}
\]

Therefore in both cases we obtain
\[
  r < \frac{H}{1 + a}
\quad\Longleftrightarrow\quad
\frac{d_M}{d_S}
< \frac{H}{1 + \dfrac{\ln M}{\ln N + \ln S}}.
\]

Either way, the unified requirement is
\[
    \frac{d_M}{d_S}
    \;<\;
    \frac{H}{\;1 + \tfrac{\ln M}{\ln n + \ln S}\;}
  .
\]

\hfill\(\square\)

\end{document}